%% file: JAIR_Example_Template.tex
\documentclass[]{jair}

\setcopyright{cc}
\copyrightyear{2025}
\acmDOI{10.1613/jair.1.17602}

\JAIRAE{Bo Han}
\JAIRTrack{} 
\acmVolume{84}
\acmArticle{20}
\acmMonth{11}
\acmYear{2025}

\RequirePackage[
  datamodel=acmdatamodel,
  style=acmauthoryear,
  backend=biber,
  giveninits=true,
  uniquename=init
  ]{biblatex}

\usepackage{graphicx}
\usepackage{multirow}
\usepackage{amsmath}
\usepackage{mathrsfs}
\usepackage{amsfonts}
\usepackage{bbding}
\usepackage{xcolor}
\usepackage{enumerate}
\usepackage{subcaption}

\begin{document}

\title[Trustworthy Transfer Learning: A Survey]{Trustworthy Transfer Learning: A Survey}





\author{Jun Wu}
\authornote{This work was mainly completed when Jun Wu was a PhD student at University of Illinois Urbana-Champaign.}
\orcid{0000-0002-1512-524X}
\email{wujun4@msu.edu}
\affiliation{%
  \institution{Michigan State University}
  \city{East Lansing}
  \state{MI}
  \country{USA}
}

\author{Jingrui He}
\orcid{0000-0002-6429-6272}
\email{jingrui@illinois.edu}
\affiliation{%
  \institution{University of Illinois Urbana-Champaign}
  \city{Urbana}
  \state{IL}
  \country{USA}}

\renewcommand{\shortauthors}{Wu \& He}

\begin{abstract}
Transfer learning aims to transfer knowledge or information from a source domain to a relevant target domain. In this paper, we understand transfer learning from the perspectives of knowledge transferability and trustworthiness. This involves two research questions: How is knowledge transferability quantitatively measured and enhanced across domains? Can we trust the transferred knowledge in the transfer learning process? To answer these questions, this paper provides a comprehensive review of trustworthy transfer learning from various aspects, including problem definitions, theoretical analysis, empirical algorithms, and real-world applications. Specifically, we summarize recent theories and algorithms for understanding knowledge transferability under (within-domain) IID and non-IID assumptions. In addition to knowledge transferability, we review the impact of trustworthiness on transfer learning, e.g., whether the transferred knowledge is adversarially robust or algorithmically fair, how to transfer the knowledge under privacy-preserving constraints, etc. Beyond discussing the current advancements, we highlight the open questions and future directions for understanding transfer learning in a reliable and trustworthy manner.

\end{abstract}



\received{12 November 2024}
\received[accepted]{12 June 2025}

\maketitle

\input{content/1_intro}

\input{content/2_prelim}

\input{content/3_transferability}

\input{content/4_trustworthiness}

\input{content/5_applications}

\input{content/6_future}

\input{content/7_conclusion}

\begin{acks}
This work is supported by National Science Foundation under Award No. IIS-2117902, and Agriculture and Food Research Initiative (AFRI) grant no. 2020-67021-32799/project accession no.1024178 from the USDA National Institute of Food and Agriculture. The views and conclusions are those of the authors and should not be interpreted as representing the official policies of the funding agencies or the government.
\end{acks}

\printbibliography

\end{document}

%% file: content/1_intro.tex
\section{Introduction}
Standard machine learning assumes that training and testing samples are independently and identically drawn (IID). With this IID assumption, modern machine learning models (e.g., deep neural networks~\cite{lecun2015deep}) have achieved promising performance in a variety of high-impact applications. However, this IID assumption is often violated in real-world scenarios, especially when samples are collected from different sources and environments~\cite{TL_survey_2010,wu2024distributional}. Transfer learning has been introduced to tackle the distribution shifts between training (source domain) and testing (target domain) data sets. In contrast to standard machine learning involving samples from a single domain, transfer learning focuses on modeling heterogeneous data collected from different domains. The intuition behind transfer learning is to bridge the gap between source and target data by discovering and transferring their shared knowledge~\cite{TL_survey_2010}. Compared to learning from the target domain alone, the transferred knowledge could significantly improve the prediction performance on the target domain, especially when the target domain has limited or no labeled data~\cite{A_distance,task_diversity}. In recent decades, by instantiating the learning models with modern neural networks, a deep transfer learning paradigm has been introduced with enhanced transferability capabilities~\cite{YosinskiCBL14}.

As illustrated in~\cite{TL_survey_2010}, transfer learning is a general term to describe the transfer of knowledge or information from source to target domains. Depending on the data and model assumptions, it can lead to various specific problem settings, such as data-level knowledge transfer (domain adaptation~\cite{A_distance,DANN,discrepancy_distance}, out-of-distribution generalization~\cite{BlanchardLS11,MuandetBS13}, and self-taught learning~\cite{RainaBLPN07}) with available source samples and model-level knowledge transfer (fine-tuning~\cite{ShachafBG21}, source-free adaptation~\cite{SHOT,AghbalouS23}, knowledge distillation~\cite{hinton2015distilling}) with a pre-trained source hypothesis. The generalization performance of transfer learning techniques under various data and model assumptions has been studied over the past decades~\cite{task_diversity,0002CZG19,minami2023transfer,MohriSS19}. In addition to generalization performance, it is crucial to understand the trustworthiness~\cite{eshete2021making} of the transferred knowledge in the transfer learning process, especially in safety-critical applications such as autonomous driving and medical diagnosis. It is explained~\cite{Varshney2022} that ``trust is the relationship between a trustor and a trustee: the trustor trusts the trustee". In the context of transfer learning, the trustor can be the owners/users/regulators of either the source or the target domain. The trustee can be the transfer learning model itself, or the knowledge transferred from the source domain to the target domain. As summarized in earlier studies~\cite{eshete2021making,Varshney2022,kaurURD23}, various trustworthiness properties can encourage the ``trustor" to trust the ``trustee" in real scenarios, including adversarial robustness, privacy, fairness, transparency, etc. Therefore, in this paper, we focus on trustworthy transfer learning~\cite{wu2023trustworthy} that aims to understand transfer learning from the perspective of both knowledge transferability and knowledge trustworthiness.  

\begin{figure}
    \centering
    \includegraphics[width=0.98\linewidth]{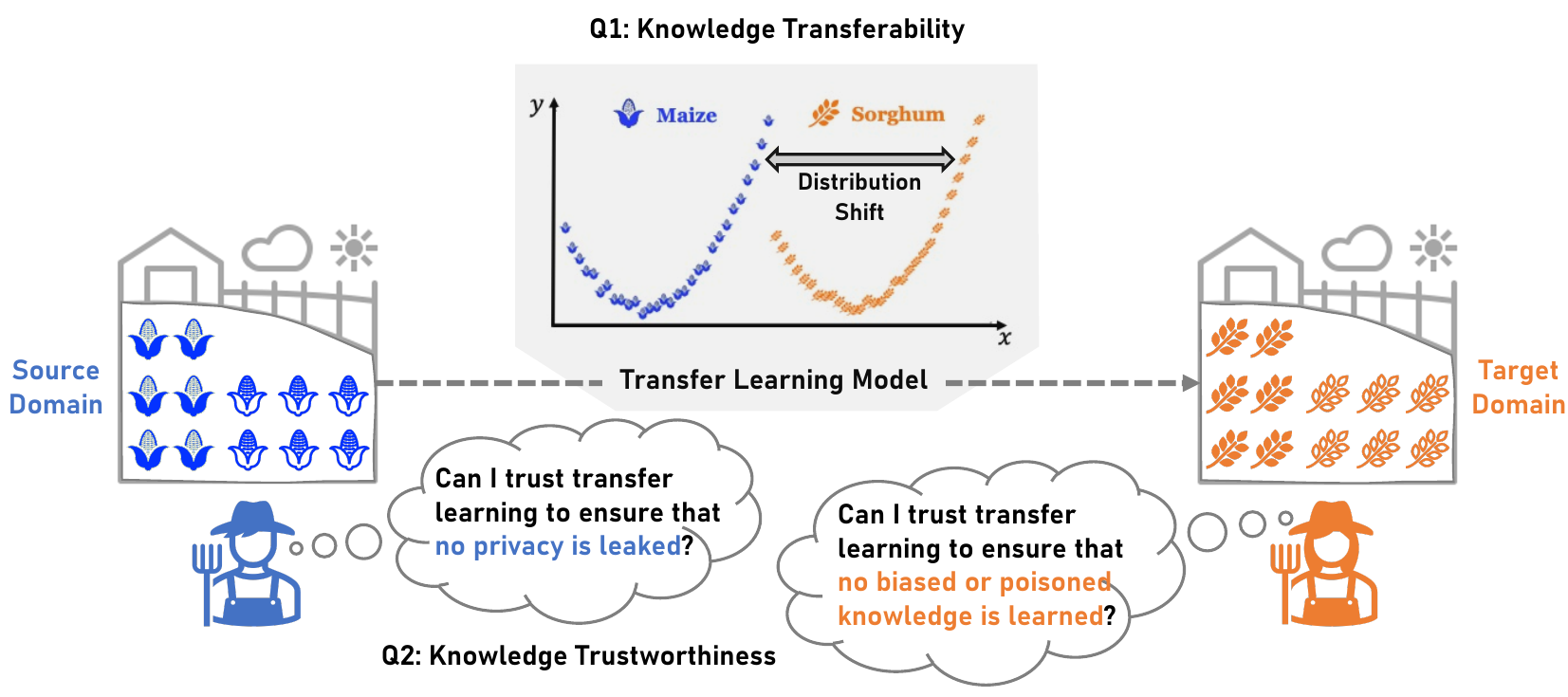}
    \vspace{-2mm}
    \caption{A motivating example of trustworthy transfer learning}
    \label{fig:motivating_example}
    \vspace{-4mm}
\end{figure}

Fig.~\ref{fig:motivating_example} provides a motivating example of trustworthy transfer learning in precision agriculture~\cite{adve2024aifarms}. In this example, a target farmer aims to train a model over the collected sorghum data. The task is to predict the biochemical traits (e.g., Nitrogen content, chlorophyll, etc.) of sorghum samples using the leaf hyperspectral reflectance~\cite{wang2023airborne,DINO}. Nevertheless, it is expensive and time-consuming to collect the labeled training samples. A feasible solution is to leverage knowledge from a relevant maize data set collected by a source farmer. This transfer learning process might involve several trustworthy concerns from the source and target farmers. To name a few, will the privacy of source data be leaked in transfer learning? How does the poisoned and biased source knowledge negatively affect the prediction performance on the target domain? What is the fundamental trade-off between transfer performance and trustworthy properties? More generally, from the perspective of data and AI model markets~\cite{pei2023data}, this emphasizes the importance of establishing trustworthiness between customers and sellers when purchasing AI models and sharing personal data.

This survey provides a comprehensive review of state-of-the-art theoretical analysis and algorithms for trustworthy transfer learning. More specifically, we summarize recent theories and algorithms for understanding knowledge transferability from two aspects: IID and non-IID transferability. IID transferability assumes that the samples within each domain are independent and identically distributed. In this scenario, we review three major quantitative metrics for evaluating the transferability across domains, including (data-level) {\em distribution discrepancy}, (task-level) {\em take diversity}, and (model-level) {\em transferability estimation}. In contrast, non-IID transferability considers a more relaxed assumption that the samples within each domain can be interdependent, e.g., connected nodes in graphs~\cite{GCN}, word occurrence in texts~\cite{LeeDS18}, temporal observations in time series~\cite{VRADA}, etc. We then review how transferability across domains can be quantitatively measured and enhanced in these complex scenarios. In addition to knowledge transferability, we also review the impact of trustworthiness on transfer learning techniques, including privacy, adversarial robustness, fairness, transparency, etc. Finally, we will highlight the open questions and future directions of trustworthy transfer learning.


The rest of this paper is organized as follows. Section~\ref{sec:preliminary} presents the main notation and the general problem definition of trustworthy transfer learning. Section~\ref{sec:transferability} and Section~\ref{sec:trustworthiness} summarize the knowledge transferability and trustworthiness in various transfer learning scenarios, respectively. Section~\ref{sec:applications} provides the applications of transfer learning techniques in real-world applications, and Section~\ref{sec:future} summarizes the open questions and future trends of trustworthy transfer learning. Finally, we conclude this survey in Section~\ref{sec:conclusion}.

%% file: content/2_prelim.tex
\section{Preliminaries}\label{sec:preliminary}
In this section, we provide the main notation and general problem definition of trustworthy transfer learning.

\begin{figure}
    \centering
    \includegraphics[width=\linewidth]{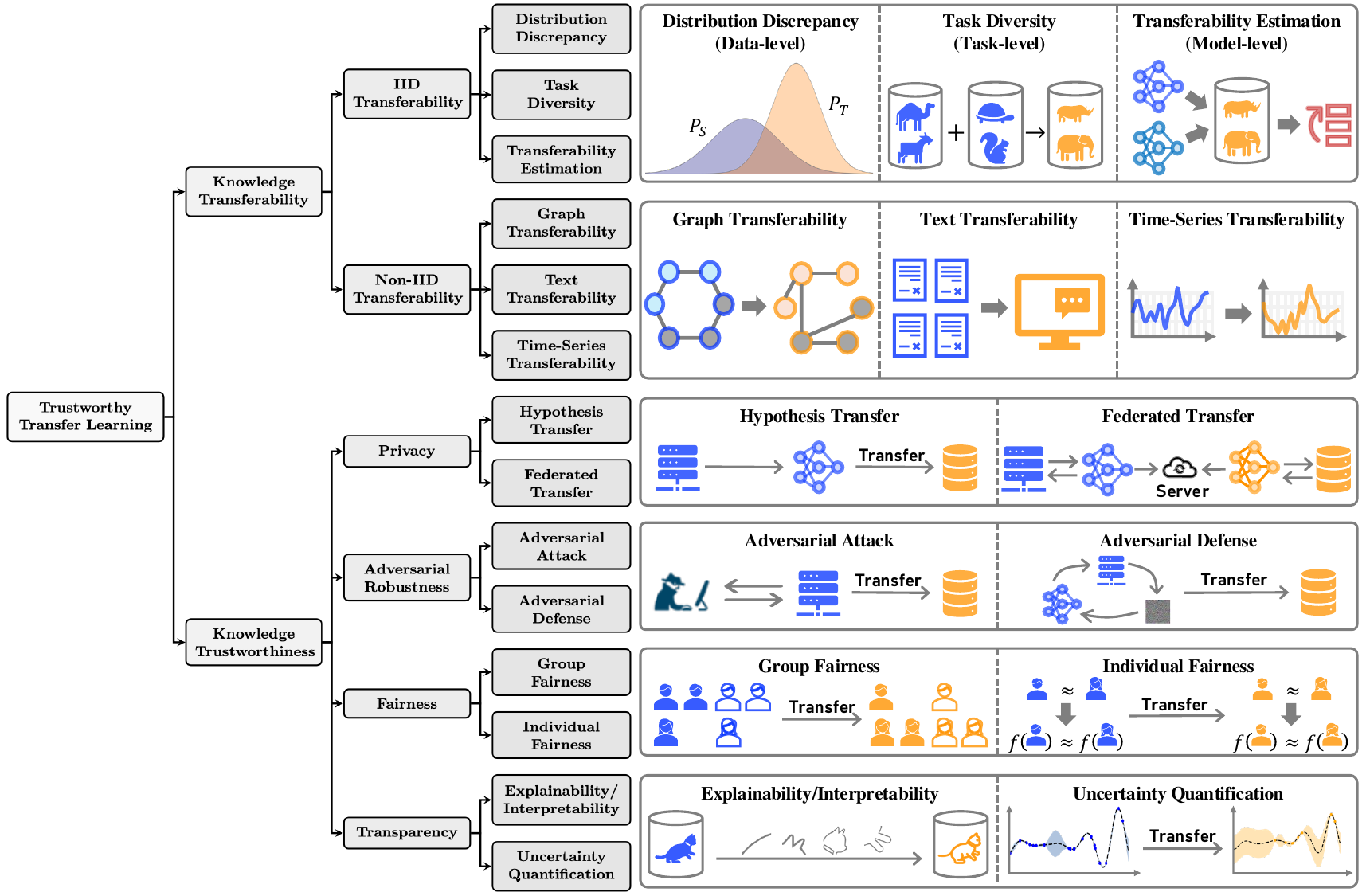}
    \caption{Overview of trustworthy transfer learning (best viewed in color)}
    \label{fig:taxonomy}
    \vspace{-3mm}
\end{figure}

\subsection{Notation}
In the paper, we let $\mathcal{X}$ and $\mathcal{Y}$ denote the input space and output space, respectively. Given a source domain $\mathcal{D}_S$ and a target domain $\mathcal{D}_T$, we denote the probability density (or mass) functions of the source and target domains as $p_S $ and $p_T$ (or $\mathrm{P}_S $ and $\mathrm{P}_T$) over $\mathcal{X}\times \mathcal{Y}$, respectively. In the context of deep transfer learning, a hypothesis function $f: \mathcal{X}\to \mathcal{Y}$ can often be decomposed into two components: a feature extraction function $g: \mathcal{X}\to \mathbb{R}^d$ and a prediction function $h: \mathbb{R}^d \to \mathcal{Y}$. We let $\mathcal{F}$ be the class of hypothesis functions (with $f\in \mathcal{F}$). Similarly, we can define $\mathcal{G}$ (with $g\in \mathcal{G}$) and $\mathcal{H}$ (with $h\in \mathcal{H}$) for the classes of the feature extraction and prediction functions, respectively. Notice that when the feature extractor is not considered (e.g., in Subsection~\ref{sec:distribution_discrepancy}), we can simply use $\mathcal{H}$ to represent the class of hypothesis functions with $h:\mathcal{X}\to \mathcal{Y}$ for any $h\in \mathcal{H}$. In addition, for any hypothesis function $f\in \mathcal{F}$ and loss function $\ell: \mathcal{Y}\times \mathcal{Y} \to \mathbb{R}$, the expected prediction errors for the source and target domains are denoted by $\mathcal{E}_S=\mathbb{E}_{\mathrm{P}_S}[\ell(f(x),y)]$ and $\mathcal{E}_T=\mathbb{E}_{\mathrm{P}_T}[\ell(f(x),y)]$, respectively.

\subsection{Problem Definition}
Transfer learning~\cite{TL_survey_2010} refers to the knowledge or information transfer from the source domain to the target domain such that the prediction performance on the target domain could be significantly improved as compared to learning from the target domain alone. Moreover, in the following definition, we generalize standard transfer learning~\cite{TL_survey_2010} to trustworthy transfer learning~\cite{wu2023trustworthy}.
\begin{definition}[Trustworthy Transfer Learning] 
Given a source domain $\mathcal{D}_S$ and a target domain $\mathcal{D}_T$, trustworthy transfer learning aims at improving the generalization and trustworthiness of a learning algorithm $f(\cdot)$ on the target domain, by leveraging latent knowledge from the source domain.
\end{definition}

The source and target domains might involve different learning tasks~\cite{TL_survey_2010,task_diversity} or data modalities~\cite{ShenLDSKNT23,IGLUE}. There are two key components in trustworthy transfer learning: knowledge transferability and trustworthiness. Specifically, knowledge transferability measures how the source knowledge can be successfully transferred to the target domain. In contrast, knowledge trustworthiness aims to answer whether transfer learning techniques provide reliable and trustworthy results in the target domain. Fig.~\ref{fig:taxonomy} provides a brief summarization of trustworthy transfer learning regarding knowledge transferability and trustworthiness (discussed in Section~\ref{sec:transferability} and Section~\ref{sec:trustworthiness}).

%% file: content/3_transferability.tex
\section{Knowledge Transferability}\label{sec:transferability}
This section summarizes the knowledge transferability in different scenarios.

\subsection{IID Transferability}
Here, we summarize different transferability indicators, including distribution discrepancy, task diversity, and transferability measures.

\subsubsection{Distribution Discrepancy}\label{sec:distribution_discrepancy}
Distribution discrepancy quantitatively measures the distribution shifts between two domains in the distribution space, when the source and target domains share the same input and output spaces (this scenario is also known as domain adaptation~\cite{TL_survey_2010}). There are different types of distribution shifts~\cite{wiles2022a}, including covariate shift~\cite{shimodaira2000improving} (feedback covariate shift~\cite{fannjiang2022conformal,prinster2023jaws}), label/target shift~\cite{lipton2018detecting,zhang2013domain}, concept shift~\cite{redko2019advances}, etc. The covariate shift holds that the conditional probability $p(y|x)$ is shared across domains, but the marginal $p(x)$ is different. Label shift assumes that the conditional probability $p(x|y)$ is shared across domains, while the marginal label distribution $p(y)$ changes. Concept shift involves changes in the conditional probability $p(x|y)$ (or $p(y|x)$), while the marginal distribution $p(y)$ (or $p(x)$) is fixed.
The integral probability metric (IPM)~\cite{muller1997integral,sriperumbudur2010hilbert,zhang2012generalization} is a general framework for quantifying the difference between two distributions, and it can be instantiated by various statistical discrepancy measures~\cite{sriperumbudur2010hilbert}, e.g., total variation distance, Wasserstein distance, maximum mean discrepancy~\cite{MMD}, etc.

\begin{definition}[Integral Probability Metric~\cite{muller1997integral}] Let $\mathrm{P}_S$ and $\mathrm{P}_T$ be the probability distributions of the source $\mathcal{D}_S$ and target $\mathcal{D}_T$ domains, respectively. The integral probability metric between $\mathrm{P}_S$ and $\mathrm{P}_T$ is defined as: 
    \begin{equation}
        d_{\mathrm{IPM}}(\mathcal{D}_S, \mathcal{D}_T) = \sup_{h\in \mathcal{H}} \left| \int_M h~ d \mathrm{P}_S - \int_M h~ d \mathrm{P}_T \right|
    \end{equation}
    where $M$ is a measurable space and $\mathcal{H}$ is a class of real-valued bounded measurable functions on $M$.
\end{definition}

The concept of distribution discrepancy is the key to theoretically understanding how knowledge can be transferred from source to target domains. For example, the seminal work of \citet{A_distance} derives a generalization bound for domain adaptation using a tractable $\mathcal{H}\Delta \mathcal{H}$-divergence. Many follow-up works have developed refined generalization bounds by introducing various discrepancy measures. The following theorem provides a unified view of such generalization bounds based on a notion of discrepancy $d(\mathcal{D}_S, \mathcal{D}_T)$.

\begin{theorem}[Unified Generalization Bound] 
Let $\mathcal{H}$ denote the hypothesis space, and $\mathcal{E}_S(h), \mathcal{E}_T(h)$ be the expected prediction error of a hypothesis $h\in \mathcal{H}$ on the source and target domains, respectively. $d(\cdot, \cdot)$ measures the difference between source and target distribution probabilities (see more instantiations below). Then for any hypothesis $h\in \mathcal{H}$, we can have a unified view of generalization error in the target domain:
\begin{align*}
    \mathcal{E}_T(h) &\leq \mathcal{E}_S(h) + d(\mathcal{D}_S, \mathcal{D}_T) + \Omega
\end{align*}
where $\Omega$ represents the redundant terms (depending on how $d(\mathcal{D}_S, \mathcal{D}_T)$ is instantiated), e.g., the difference of labeling functions across domains~\cite{A_distance,f_divergence}, the complexity of hypothesis space $\mathcal{H}$~\cite{discrepancy_distance,zhang2012generalization}, number of training samples~\cite{A_distance,redko2017theoretical}, etc.
\end{theorem}
We have the following observations regarding this unified view of generalization error. (1) The complexity of the class of hypothesis functions $\mathcal{H}$ plays a crucial role in deriving tight generalization error bounds. Various metrics have been applied to quantify this complexity~\cite{redko2019advances}, including the Vapnik-Chervonenkis (VC) dimension~\cite{ben2006analysis,blitzer2007learning,A_distance,peng2019moment}, Rademacher complexities~\cite{f_divergence,SCA,MDD,discrepancy_distance,mohri2012new}, covering number~\cite{MDD,zhang2012generalization}, etc. We refer the reader to the survey~\cite{redko2019advances} for more discussion. (2) Generally, the discrepancy $d(\mathcal{D}_S, \mathcal{D}_T)$ measures the difference between source and target distributions over the joint space $\mathcal{X}\times \mathcal{Y}$, when the distribution shifts occur across domains. In practice, it is commonly seen that the discrepancy $d(\mathcal{D}_S, \mathcal{D}_T)$ is defined over the input space $\mathcal{X}$ (when no label information is available in unsupervised domain adaptation), or over the joint space $\mathcal{X}\times \mathcal{Y}$ (when (pseudo)labels for target samples are available). The first type of $d(\mathcal{D}_S, \mathcal{D}_T)$, defined over the input space $\mathcal{X}$, is often associated with the covariate shift assumption~\cite{shimodaira2000improving} or the redundant term indicating the difference of labeling functions across domains, when deriving the generalization error bound. In the following, we summarize several commonly used discrepancy metrics $d(\mathcal{D}_S, \mathcal{D}_T)$.

\begin{itemize}
    \item {\em Total Variation Distance:} The total variation distance (also referred to as $L^1$ divergence)~\cite{ben2006analysis} between source and target domains can be defined as
    \begin{equation}\label{eq:tv}
        d_{\mathrm{TV}}(\mathcal{D}_S, \mathcal{D}_T) = \sup_{B\in \mathcal{B}} \left| \mathrm{P}_S [B] - \mathrm{P}_T [B] \right| 
    \end{equation}
    where $\mathcal{B}$ is the set of measurable subsets under $\mathrm{P}_S$ and $\mathrm{P}_T$. It is shown~\cite{sriperumbudur2010hilbert} that the total variation distance can be considered as a special case of the integral probability metric.
    
    \item {\em $\mathcal{H}\Delta\mathcal{H}$-divergence:} It is illustrated~\cite{ben2006analysis,A_distance} that the empirical estimate of the total variation distance in Eq. (\ref{eq:tv}) has two limitations. First,  it cannot be accurately estimated from finite samples of arbitrary distributions in practice. Second, it results in loose generalization bounds due to involving a supremum over all measurable subsets. To address these limitations, \citet{ben2006analysis,blitzer2007learning} introduce the following $\mathcal{H}\Delta\mathcal{H}$-divergence:
    \begin{equation}
        d_{\mathcal{H}\Delta\mathcal{H}}(\mathcal{D}_S, \mathcal{D}_T) = \sup_{h, h'\in \mathcal{H}} \left| \mathrm{P}_S \left[ h(x) \neq h'(x) \right] - \mathrm{P}_T \left[ h(x) \neq h'(x) \right] \right|
    \end{equation}
    It can be seen that this distribution difference is defined over the hypothesis-dependent subsets $\{ \mathbb{I}[h(x) \neq h'(x)] | h, h' \in \mathcal{H} \}$.

    \item {\em Discrepancy Distance:} \citet{discrepancy_distance} extends the $\mathcal{H}\Delta\mathcal{H}$-divergence to a more general discrepancy distance for measuring distribution differences.
    \begin{equation}\label{eq:disc}
        d_{\mathrm{disc}}(\mathcal{D}_S, \mathcal{D}_T) = \max_{h, h'\in \mathcal{H}} \left| \mathbb{E}_{x\sim \mathrm{P}_S}\left[ \ell\left( h(x), h'(x) \right) \right] - \mathbb{E}_{x\sim \mathrm{P}_T}\left[ \ell\left( h(x), h'(x) \right) \right] \right| 
    \end{equation}
    where $\ell(\cdot, \cdot)$ denotes a general loss function (though the derived generalization bounds require that $\ell(\cdot, \cdot)$ is symmetric and obeys the triangle inequality). When using 0-1 classification loss, this discrepancy distance exactly recovers the $\mathcal{H}\Delta\mathcal{H}$-divergence. The discrepancy distance can be flexibly applied to compare distributions across various tasks, e.g., regression~\cite{discrepancy_distance,cortes2011domain}.

    \item $\mathcal{Y}$-discrepancy: It is notable that the discrepancy distance in Eq. (\ref{eq:disc}) quantifies the difference between two marginal distributions over $\mathcal{X}$, when the ground-truth labeling function is unknown in the target domain. Later, \citet{mohri2012new} further extend the discrepancy distance to the $\mathcal{Y}$-discrepancy, which is defined over $\mathcal{X}\times \mathcal{Y}$ as follows.
    \begin{equation}
        d_{\mathcal{Y}}(\mathcal{D}_S, \mathcal{D}_T) = \sup_{h\in \mathcal{H}} \left| \mathbb{E}_{(x, y)\sim \mathrm{P}_S}\left[ \ell\left( h(x), y \right) \right] - \mathbb{E}_{(x, y)\sim \mathrm{P}_T}\left[ \ell\left( h(x), y \right) \right] \right|
    \end{equation}
    In practice, this discrepancy can be estimated using pseudo labels of the target data, when there are no labeled data in the target domain~\cite{long2018conditional,courty2017joint}.

    \item Margin Disparity Discrepancy: \citet{MDD} extend the notion of discrepancy distance in Eq. (\ref{eq:disc}) to margin disparity discrepancy (MDD) in multi-class classification settings. Specifically, MDD involves two key refinements: (1) the use of a margin-based loss function, and (2) the formulation of the discrepancy over both a hypothesis space $\mathcal{H}$ and a specific classifier $h$.
    \begin{equation}
        d_{\mathrm{MDD}}(\mathcal{D}_S, \mathcal{D}_T) = \sup_{h'\in \mathcal{H}} \left| \mathbb{E}_{x\sim \mathrm{P}_S}\left[ \Phi_{\rho} \left( \rho_{h'}(x, h(x)) \right) \right] - \mathbb{E}_{x\sim \mathrm{P}_T}\left[ \Phi_{\rho} \left( \rho_{h'}(x, h(x)) \right) \right] \right|
    \end{equation}
    where the function $\rho_{h'}(\cdot, \cdot)$ defines the margin of a hypothesis $h'$, and the function $\Phi_{\rho}(\cdot)$ defines the margin-based loss over a threshold $\rho>0$. 

    \item $f$-divergence: Building on the margin disparity discrepancy (MDD)~\cite{MDD}, \citet{f_divergence} further develop a generic notion of the discrepancy based on the variational characterization of $f$-divergence~\cite{nguyen2010estimating}. Specifically, the $f$-divergence is bounded by
    \begin{equation*}
        d_{f}(\mathcal{D}_S, \mathcal{D}_T) = \int p_T(x) \phi \left( \frac{p_S(x)}{p_T(x)} \right) dx \geq \sup_{T\in \mathcal{T}} \mathbb{E}_{x\sim \mathrm{P}_S}\left[ T\left( x \right) \right] - \mathbb{E}_{x\sim \mathrm{P}_T}\left[ \phi^* \left( T\left( x \right) \right) \right]
    \end{equation*}
    where $\phi(\cdot)$ is a convex lower semi-continuous function that satisfies $\phi(1)=0$. $\phi^*$ is the conjugate function of $\phi$. $\mathcal{T}$ is a set of measurable functions. Based on this observation, ~\cite{f_divergence} define a notion of $f$-divergence guided discrepancy as follows.
    \begin{equation}\label{eq:f_divergence}
        d_{\phi}(\mathcal{D}_S, \mathcal{D}_T) = \sup_{h'\in \mathcal{H}} \left| \mathbb{E}_{x\sim \mathrm{P}_S}\left[ \ell\left( h(x), h'(x) \right) \right] - \mathbb{E}_{x\sim \mathrm{P}_T}\left[ \phi^* \left( \ell\left( h(x), h'(x) \right) \right) \right] \right| 
    \end{equation}
    The flexibility in choosing $\phi^*$ enables $f$-divergence to recover many popular statistical divergences, e.g., Jensen-Shannon (JS) divergence, Kullback-Leibler (KL) divergence, Reverse KL (KL-rev) divergence, Pearson $\chi^2$ divergence, etc. As a result, different choices of $\phi^*$ can define various discrepancies from Eq. (\ref{eq:f_divergence}). Besides, it is shown that the notion of discrepancy in Eq. (\ref{eq:f_divergence}) can also recover MDD.

    \item Generalized Discrepancy: In contrast, \citet{generalized_discrepancy,cortes2019adaptation} generalize the discrepancy distance in Eq. (\ref{eq:disc}) using reweighting techniques. That is, the difference between two distributions can be adjusted by multiplying the loss for each training example by a non-negative weight~\cite{huang2006correcting,cortes2008sample,zhang2013domain}. Formally, for any hypothesis-dependent reweighting function $U_h$, the generalized discrepancy is defined as follows.
    \begin{equation}
        d_{\mathrm{DISC}}(\mathcal{D}_S, \mathcal{D}_T) = \max_{h\in \mathcal{H}, h''\in \mathcal{H}'' } \left| \mathbb{E}_{x\sim \hat{\mathrm{P}}_S}\left[ {U}_h(x) \cdot \ell\left( h\left(x\right), f_S\left(x\right) \right) \right] - \mathbb{E}_{x\sim \hat{\mathrm{P}}_T}\left[ \ell\left( h(x), h''(x) \right) \right] \right|
    \end{equation}
    where $\hat{\mathrm{P}}_S, \hat{\mathrm{P}}_T$ denote the empirical distributions of source and target domains, respectively, and $f_S$ denotes the source labeling function. 

    \item R{\'e}nyi Divergence: Furthermore, \citet{MansourMR09,cortes2010learning} derive the generalization bounds for adaptation approaches based on importance reweighting (e.g., sample reweighting for single-source adaptation~\cite{cortes2010learning} and domain reweighting for multi-source adaptation~\cite{MansourMR09}) based on the following R{\'e}nyi divergence~\cite{renyi1961measures}.
    \begin{equation}
        d_{\alpha}(\mathcal{D}_T || \mathcal{D}_S) = \frac{1}{\alpha - 1} \log \sum_{x\in \mathcal{X}} \mathrm{P}_T (x) \left( \frac{\mathrm{P}_T (x) }{ \mathrm{P}_S (x) } \right)^{\alpha - 1}
    \end{equation}
    where $\alpha \geq 0$.

    \item Wasserstein Distance~\cite{Wasserstein}: In general, fro any $p\geq 1$, the $p$-Wasserstein distance between two distributions can be defined as follows.
    \begin{equation}
        d_{W_p}(\mathcal{D}_S, \mathcal{D}_T) = \left( \inf_{\pi \in \Pi (\mathrm{P}_S, \mathrm{P}_T) } \int c(x, x')^p d \pi(x, x') \right)^{1/p}
    \end{equation}
    where $\Pi (\mathrm{P}_S, \mathrm{P}_T)$ is the set of all measures over $\mathcal{X}\times \mathcal{X}$ with marginals $\mathrm{P}_S$ and $\mathrm{P}_T$, and $c(\cdot, \cdot)$ is a distance function. When $p=1$, the 1-Wasserstein distance (also known as earth mover's distance) is one special case of the integral probability metric with $\mathcal{H} = \{h: ||h||_L \leq 1 \}$. 
    Specifically, based on the Kantorovich-Rubinstein duality~\cite{dudley2002real}, it holds that
    \begin{equation}
        d_{W_1}(\mathcal{D}_S, \mathcal{D}_T) = \inf_{\pi \in \Pi (\mathrm{P}_S, \mathrm{P}_T) } \int c(x, x') d \pi(x, x') = \sup_{||h||_L\leq 1} \mathbb{E}_{\mathrm{P}_S}[h(x)] - \mathbb{E}_{\mathrm{P}_T}[h(x)]
    \end{equation}
    where $||h||_L = \sup_{x\neq x'} |h(x) - h(x')|/c(x, x')$. This enables a practical empirical estimation of the 1-Wasserstein distance using gradient descent optimization~\cite{arjovsky2017wasserstein}. Therefore, \citet{Wasserstein,redko2017theoretical} apply the 1-Wasserstein distance to analyze distribution shifts between source and target domains.

    \item Maximum Mean Discrepancy: \citet{tzeng2014deep,long2015learning} leverage the maximum discrepancy discrepancy (MMD) \cite{MMD} to measure the distribution difference between source and target domains. MMD can be considered as another special case of the integral probability metric by instantiating the hypothesis space with a unit ball in a reproducing kernel Hilbert space associated with kernel $k(\cdot, \cdot)$. Given a kernel function $k(\cdot, \cdot)$, the MMD between source and target distribution can be defined as:
    \begin{equation}
        \begin{aligned}
            d_{\mathrm{MMD}}(\mathcal{D}_S, \mathcal{D}_T) &= \mathbb{E}_{x_S, x'_S \sim \mathrm{P}_S} \big[ k\left(x_S, x'_S \right) \big] - 2 \mathbb{E}_{x_S\sim \mathrm{P}_S, x_T\sim \mathrm{P}_T } \big[ k\left(x_S, x_T \right) \big] \\
            &\quad + \mathbb{E}_{x_T, x'_T\sim \mathrm{P}_T} \big[ k\left(x_T, x'_T \right) \big]
        \end{aligned}
    \end{equation}
    \citet{redko2019advances} further show the generalization error bounds based on MMD.
    
    \item Cauchy-Schwarz Divergence~\cite{yin2024domain}: Recently, \citet{yin2024domain} use the Cauchy-Schwarz (CS) divergence~\cite{principe2010information} to theoretically understand the knowledge transferability across domains.
    \begin{equation}
        d_{\mathrm{CS}}(\mathcal{D}_S, \mathcal{D}_T) = - \log \left( \frac{\left( \int p_S (x) p_T (x) dx \right)^2 }{\int p_S^2 (x) dx \cdot \int p_T^2 (x) dx } \right)
    \end{equation}
    It is shown~\cite{yin2024domain} that this CS divergence can lead to tighter generalization error bounds than the KL divergence~\cite{nguyen2022kl}. Besides, the empirical estimate of the CS divergence is closely related to MMD~\cite{MMD}.
\end{itemize}
From the perspective of empirical estimation, the discrepancy measures can be broadly categorized into two groups. The first group includes statistical discrepancy measures~\cite{nguyen2022kl,CORAL,RSD}, such as Maximum Mean Discrepancy (MMD)~\cite{tzeng2014deep,long2015learning} and Wasserstein distance~\cite{courty2016optimal,redko2017theoretical,fatras2021unbalanced}, which can be directly estimated from finite samples. The second group is based on adversarial learning~\cite{DANN,saito2018maximum,tzeng2017adversarial,hoffman2018cycada,MDD,f_divergence}, which requires an additional neural network to optimize an adversarial objective. More recently, \citet{KashyapHKZ21,YuanHWMLSH22} provide empirical comparisons of various discrepancy measures in natural language processing and computer vision tasks. It is noted that when there are no labeled samples in the target domain, one common strategy in designing practical domain adaptation algorithms is to minimize the discrepancy across domains over $\mathcal{X}$. However, it has been shown~\cite{Ben-DavidLLP10,0002CZG19,WuWKL19,JohanssonSR19} that exact marginal distribution matching might lead to negative transfer in practice.

The notion of distribution discrepancy has been applied to understand knowledge transferability in various realistic adaptation scenarios, including single-source adaptation~\cite{cortes2010learning,A_distance,MDD,f_divergence,nguyen2022kl}, multi-source adaptation~\cite{mansour2021theory,hoffman2018algorithms,wu2024distributional}, open-set adaptation~\cite{fang2020open,RAINCOAT}, domain generalization~\cite{MuandetBS13,BlanchardLS11} (also known as out-of-distribution generalization), privacy-preserving federated adaptation~\cite{PengHZS20}, dynamic adaptation~\cite{KumarML20,WuH22}, etc.

\subsubsection{Task Diversity}\label{sec:task_diversity}
Task diversity~\cite{task_diversity,WatkinsUNA23} is another tool for theoretically understanding the performance of transfer learning. It enables a relaxed data assumption that the source and target domains can have different output spaces, i.e., each domain can be associated with a different learning task~\cite{TL_survey_2010}. In the context of transfer learning, it assumes that a generic nonlinear feature representation function is shared across all tasks. Then each task is associated with a shared representation learning function and a task-specific prediction function. In~\cite{task_diversity}, task diversity is defined to characterize how the worst-case representation difference can be controlled when the task-averaged representation difference is small. In this case, the worst-case representation difference is the distance between two representation functions with the worst-case task-specific prediction function, while the task-averaged representation difference indicates the distance between two representation functions over all the training tasks. 
\begin{definition}[Task Diversity~\cite{task_diversity}]
    Given $N$ source tasks associated with a representation function class $\mathcal{H}$ and a prediction function class $\mathcal{H}$, let $h_i \in \mathcal{H}$ represent the task-specific prediction function for the $i^{\mathrm{th}}$ source task ($i=1,2\cdots,N$), we say that source tasks with the functions $\{h_1,h_2, \cdots, h_N\}$ are $\left( \nu, \epsilon \right)$-diverse over the function class $\mathcal{H}_0$ for a representation function $g \in \mathcal{G}$, if uniformly for all $g' \in \mathcal{G}$,
    \begin{equation}
        \underbrace{ \sup_{ h_0 \in \mathcal{H}_0 } \inf_{h'\in \mathcal{H}} \left\{ \mathcal{E}_{T} \left( h' \circ g' \right) - \mathcal{E}_{T} \left( h_0 \circ g \right) \right\} }_{\text{Worst-case representation difference}} \leq \frac{1}{\nu}\cdot \underbrace{ \left( \frac{1}{N}\sum_{i=1}^N \inf_{h'\in \mathcal{H}} \left\{ \mathcal{E}_{S_i} \left( h' \circ g' \right) - \mathcal{E}_{S_i} \left( h_i \circ g \right) \right\} \right) }_{\text{Task-averaged representation difference}} ~~ + ~~~ \epsilon
    \end{equation}
    where $\mathcal{E}_T(h\circ g)$ represents the expected error in the target task using a representation learning function $g$ and a prediction function $h$, and $\mathcal{E}_{S_i}(h\circ g)$ represents the expected error in the $i^{\mathrm{th}}$ source task.
\end{definition}
Based on the task diversity, \citet{task_diversity} derive the excess risk bounds of transfer learning for the target task in terms of the complexity of the shared representation function class $\mathcal{G}$, the complexity of the prediction function class $\mathcal{H}$, the number of tasks $N$, and the number of training samples for each task (both source and target). Furthermore, \citet{WatkinsUNA23} show that under the Lipschitz assumption for the loss function, the excess risk in the target task only achieves the standard rate of $\mathcal{O}(n_T^{-1/2})$, where $n_T$ is the number of training samples in the target task. By using the smoothness assumption for the loss function~\cite{srebro2010optimistic}, they derive optimistic rates that interpolate between the standard rate of $\mathcal{O}(n_T^{-1/2})$ and the fast rate of $\mathcal{O}(n_T^{-1})$ for the excess risk in the target task.

In addition, \citet{du2021fewshot,tripuraneni2021provable} consider a simplified version of task diversity in the cases of linear prediction functions and quadratic loss. They also theoretically show the benefits (i.e., reduced sample complexity in the target task induced by all available source samples) of representation learning from source tasks. Based on the task diversity, \citet{xu2021representation} analyze more realistic learning scenarios in which the source and target tasks use different prediction function spaces. However, all the aforementioned theoretical analyses assume uniform sampling from each source task, i.e., all source tasks are equally important for learning a representation function. Instead, \citet{chen2022active,chen2023active,wang2023improved} study active transfer learning by quantifying the task relatedness and selecting the source tasks that are most relevant to the target task. Similarly, \citet{xu2024towards} explore the selection of source tasks for multi-task fine-tuning of foundation models, e.g., fine-tuning the foundation model on auxiliary source tasks before adapting it to the target task with limited labeled samples. More recently, \citet{zhao2023blessing} show that pre-training on a single source task with a high diversity of classes can provably improve the sample efficiency of the downstream tasks. In contrast, \citet{cole2024provable} leverage task diversity to understand the in-context learning behavior of foundation models.

\begin{figure}[!t]
  \centering
  \begin{subfigure}{\textwidth}
    \centering
    \includegraphics[width=\textwidth]{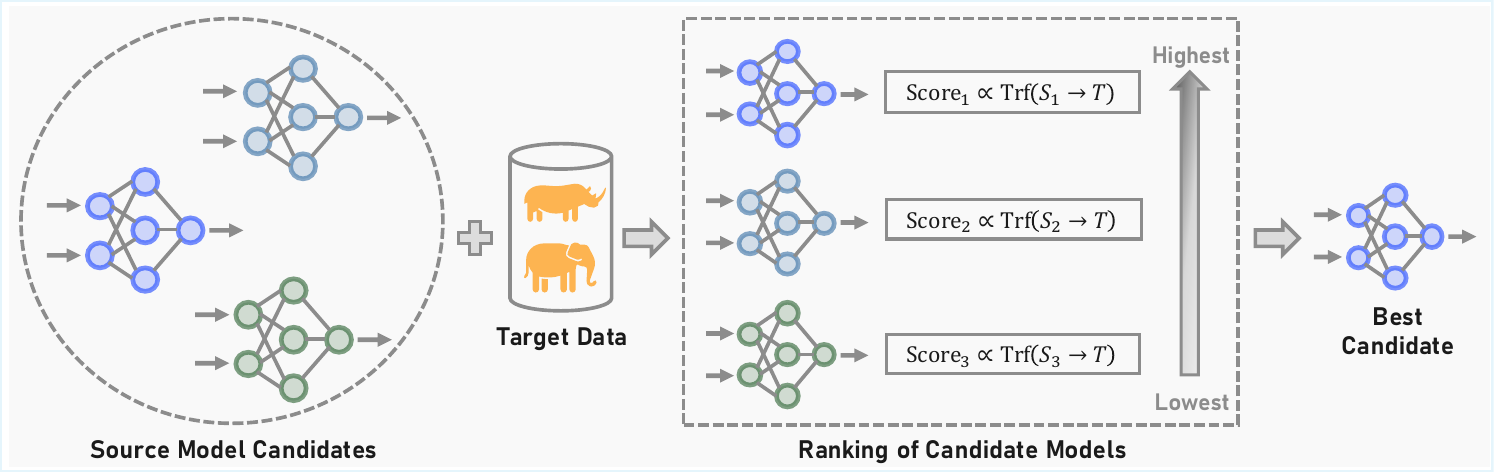}
    \caption{Transferability estimation for selecting the source model}
    \label{fig:transferability_metric_a}
  \end{subfigure}
  
  \vspace{0.75em}
  
  \begin{subfigure}{\textwidth}
    \centering
    \includegraphics[width=\textwidth]{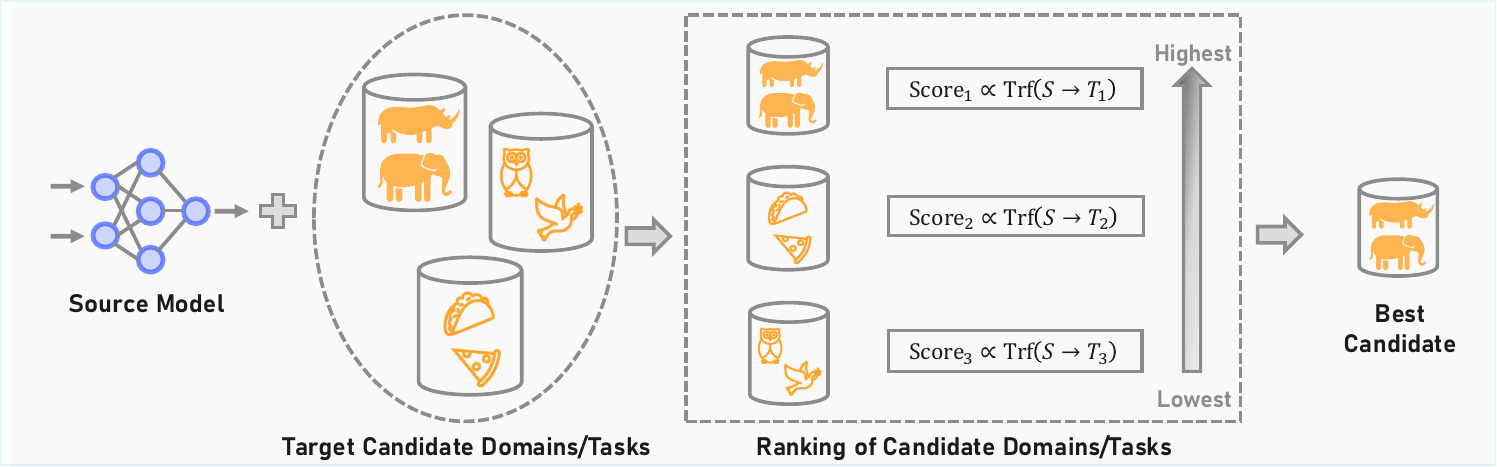}
    \caption{Transferability estimation for selecting the target domain/task}
    \label{fig:transferability_metric_b}
  \end{subfigure}
  \vspace{-7mm}
  \caption{Evaluation of transferability between the pre-trained source model and the target data: (a) Transferability scores select the best source model for the target data given a large pool of pre-trained source models. (b) Transferability scores identify the most suitable application domains/tasks for a source model.}
  \label{fig:transferability_metric}
  \vspace{-2mm}
\end{figure}

\subsubsection{Transferability Estimation}
In contrast to the data-centric transferability analyses in Subsection~\ref{sec:distribution_discrepancy} and Subsection~\ref{sec:task_diversity}, this subsection explores the knowledge transferability of pre-trained source models. This is driven by the rapidly expanding open-source model repositories such as HuggingFace~\cite{Transformers} and PyTorch Hub~\cite{PyTorch}. Fine-tuning a pre-trained source model on downstream target data sets with limited sample sizes improves model accuracy and robustness~\cite{HendrycksLM19}. A natural question arises in this scenario: {\em Given a large pool of pre-trained source models, how can we efficiently select the best one for a target data set?} As shown in Fig.~\ref{fig:transferability_metric}, another relevant question is how to identify the most suitable domains/tasks for a given pre-trained source model. One trivial solution is brute-force fine-tuning, where all source models are fine-tuned individually and then ranked based on their transfer accuracy. However, this method is highly time-consuming and computationally expensive. To solve this problem, transferability estimation has been studied to quantitatively measure how effectively the knowledge can be transferred from a pre-trained source model to a target domain~\cite{H-Score,NCE,LEEP,AgostinelliPUMF22,IbrahimPM22}. Following~\cite{NCE}, given a pre-trained source model $f_S(\cdot)$, the transferability from $f_S(\cdot)$ to a target domain associated with sampling distribution $\mathrm{P}_T$ can be defined below.

\begin{definition}[Transferability Measure~\cite{NCE}]
    The transferability from a pre-trained source model $\mathcal{D}_S$ to a target domain $\mathcal{D}_T$ with sampling distribution $\mathrm{P}_T$ is measured by the expected accuracy of the fine-tuned model on the target domain:
    \begin{equation}
        \mathrm{Trf}(S \to T) = \mathbb{E}_{(x, y)\sim \mathrm{P}_T}\left[ \mathrm{acc} \left( x, y; f_T \right) \right]
    \end{equation}
    where $f_T(\cdot)$ is the fine-tuned model from a pre-trained source model $f_S(\cdot)$, and $\mathrm{acc}(\cdot)$ indicates the prediction accuracy.
\end{definition}
Thus, a good transferability measurement should have two key properties: (1) the learned transferability score correlates well with the transfer accuracy of the fine-tuned model on the target domain, and (2) it should be significantly more efficient than the fine-tuning approach. Notably, the transferability score does not exactly predict the accuracy of the fine-tuned model on the target domain in practice. Instead, it only needs to correlate with the ranking of fine-tuning accuracy among a pool of source models, i.e., a high transferability score indicates a better source model resulting in higher transfer accuracy.


The estimation of the transferability score is related to the architectures of pre-trained source models. In scenarios where the source models are trained on supervised classification tasks, every source model $f_S(\cdot)$ is associated with a feature extractor and a predictor. In this case, NCE~\cite{NCE} leverages conditional entropy to define the transferability score, assuming that source and target domains share the same input samples but different labels. It is motivated by the observation that the optimal average log-likelihood on target training samples is lower bounded by the negative conditional entropy. Similarly, LEEP~\cite{LEEP} estimates the average log-likelihood of target samples using the dummy label distributions generated from the pre-trained source model. Notably, it is shown that LEEP is an upper bound of the NCE~\cite{NCE} plus the average log-likelihood of the dummy labels. It is computationally efficient by using only a single forward pass of the source model through the target data. Nevertheless, LEEP can not handle unsupervised and self-supervised pre-trained models with only a feature extractor. 

To address this problem, recent works~\cite{PARC,LogME,NguyenTHDTHN23} have utilized only the feature extractor $f_S(\cdot)$ to define transferability scores. They can be divided into two frameworks. One is to measure the class separability of the target samples in the feature space induced by $f_S(\cdot)$. For example, H-score~\cite{H-Score} is defined based on the inter-class variance and feature redundancy of target samples learned from the pre-trained source model. It is inspired by the connection between the optimal prediction error and the modal decomposition of the divergence transition matrix~\cite{huang2024universal}. The follow-up work~\cite{IbrahimPM22} introduces a shrinkage-based H-score to improve the covariance estimation of H-score~\cite{H-Score} in high-dimensional feature spaces. $\mathcal{N}$LEEP~\cite{NLEEP} replaces the dummy label generation module of LEEP~\cite{LEEP} with a new Gaussian Mixture Model (GMM). 
Furthermore, GBC~\cite{GBC} maps each target class as a Gaussian distribution and estimates the pair-wise class separability (i.e., the amount of overlap between two class-wise Gaussian distributions) using the Bhattacharyya coefficient~\cite{bhattacharyya1946measure}.
The other one is to add a probabilistic linear transformation that maps the feature space of $f_S(\cdot)$ to the target output space in a Bayesian framework. For example, LogME~\cite{LogME} is defined over marginalized likelihood $p(y_i | x_i; f_S) = \int p(w) p(y_i | f_S(x_i), w) dw$, assuming that the prior distribution of the newly added linear transformation $w$ is an isotropic multivariate Gaussian $w \sim \mathcal{N}(0, \alpha^{-1} \mathbb{I})$. The follow-up work PACTran~\cite{PACTran} further defines a theoretically grounded family of transferability scores based on the optimal PAC-Bayesian error bound~\cite{GermainBLL16}, taking into consideration various instantiations of the prior distribution for the linear transformation $w$, such as Dirichlet, Gamma, and Gaussian priors. Additionally, TransRate~\cite{TransRate} exploits the coding rate to estimate transferability scores of any intermediate layer within the pre-trained model. More recently, inspired by neural scaling laws for fine-tuned LLMs~\cite{tay2022scale}, \citet{lin2024selecting} investigatet the transferability estimation of large language models (LLMs) based on a rectified scaling law that characterizes the connection between the fine-tuned test loss and the number of target samples.

The transferability metrics mentioned above enable the selection of the best source model from a large pool of open-sourced pre-trained models. Recent studies~\cite{SFDA,BBGNKB23} take one step further by studying source model ensemble selections, which transfer knowledge from multiple pre-trained source models to target training samples. This line of research is inspired by the success of ensemble machine learning models~\cite{Lakshminarayanan17} in improving model performance. Specifically, \citet{AgostinelliUMF22} extend LEEP~\cite{LEEP} to select source model ensembles under the assumption that the source models operate independently. In contrast, OSBORN~\cite{BBGNKB23} relaxes this assumption and explores the inter-model cohesion among source models to estimate the transferability of an ensemble of models to a target domain.

\subsection{Non-IID Transferability}
The IID assumption is often violated in real scenarios, e.g., connected nodes in graphs~\cite{GCN}, word occurrence in texts~\cite{LeeDS18}, temporal observations in time series~\cite{VRADA}, etc. To bridge this gap, non-IID transferability explores the knowledge transfer across domains, assuming that samples within each domain can be interdependent.

\begin{figure}
    \centering
    \includegraphics[width=\linewidth]{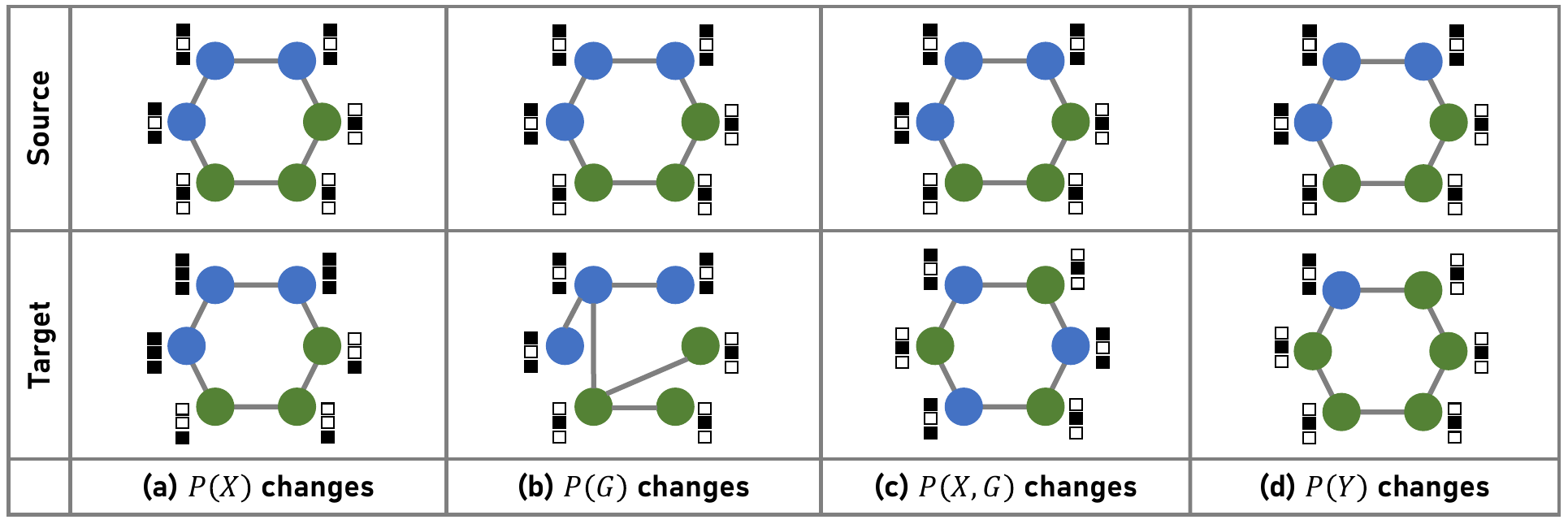}
    \vspace{-6mm}
    \caption{Illustration of distribution shifts in characterizing graph transferability. The node distribution within the graph can be represented by $\mathrm{P}(X, G, Y)$ where $X, G, Y$ denote the input node attributes, topology structure, and output class labels, respectively. The color of nodes indicates the class labels $Y$ (blue or green).}
    \label{fig:graph_transfer}
    \vspace{-3mm}
\end{figure}

\subsubsection{Transferability on Graph Data}
Graph data is being generated across a variety of application domains, ranging from bioinformatics~\cite{MPNN} to e-commerce~\cite{GRADE}, from protein-protein interaction prediction~\cite{GraphSAGE} to social network analysis~\cite{GIN}. To capture the complex structure of graph data, graph neural networks (GNNs)~\cite{ScarselliGTHM09,DefferrardBV16} have been introduced to encode the nodes within the graphs into low-dimensional vector representations. It is shown~\cite{zhang2019graph} that there are two major learning paradigms for GNNs: spectral-based~\cite{GCN,DefferrardBV16} and spatial-based GNNs~\cite{GraphSAGE,GIN,MPNN}. Recently, the transferability of spectral and spatial GNNs has been studied by exploring whether GNNs are transferable across graphs of varying sizes and topologies~\cite{WNN,GRADE}. In this survey, we focus on understanding the transferability of GNNs in node-level graph learning tasks. Note that the key challenge to theoretically understand the transferability of GNNs is to measure the distribution shifts of two graphs. As illustrated in Fig.~\ref{fig:graph_transfer}, the distribution shifts between source and target graphs are generally induced by the joint probabilities $\mathrm{P}_S(X, G, Y)$ and $\mathrm{P}_T(X, G, Y)$, where $X, G, Y$ denote the input node attributes, topology structure, and output class labels, respectively. Specifically, the transferability of spectral GNNs leverages the graph limits, e.g., graphon~\cite{WNN,maskey2023transferability} and graphop~\cite{LeJ23}, to determine if two graphs represent the same underlying structure as the number of nodes goes to infinity. In contrast, the transferability of spatial GNNs typically relies on the empirical distribution differences of node representations in a latent embedding space learned by GNNs~\cite{GRADE}.

Spectral GNNs define the graph convolutions in the spectral domain using the graph Fourier transform from the perspective of graph signal processing~\cite{DefferrardBV16,GCN}. Recent efforts~\cite{RuizGR21,levie2021transferability,maskey2023transferability,WNN,LeJ23} have been dedicated to understanding the transferability of spectral GNNs by answering the following question: {\em Can spectral GNNs trained on a source graph perform well on a target graph of different sizes?} This question is also known as size generalization~\cite{YehudaiFMCM21,BevilacquaZ021}. The intuition behind the transferability of spectral GNNs is that if two graphs represent the same underlying phenomenon, their GNN outputs will be similar. Thus, the transferability of spectral GNNs can be derived from various aspects, including generic topological space~\cite{levie2021transferability}, graphon~\cite{WNN}, graphop~\cite{LeJ23}, and $k$-hop ego-graph~\cite{EGI}. To be more specific, \citet{levie2019transferability,levie2021transferability} study the transferability of spectral graph filters on different discretizations of the same underlying continuous topological space. Later, graphon theory~\cite{lovasz2012large} is used to analyze the transferability of spectral GNNs. Formally, a graphon is defined by a bounded symmetric kernel and can be viewed as a graph with an uncountable number of nodes. In particular, \citet{WNN} leverage graphon to study the asymptotic behavior of GNNs~\cite{DefferrardBV16}, showing that GNNs converge to graphon neural networks (WNNs) as the number of nodes increases to infinity. This convergence implies that under mild assumptions, GNNs are transferable across graphs with performance guarantees if both graphs are drawn from the same graphon~\cite{WNN,RuizGR21,maskey2023transferability}. Following this observation, recent works~\cite{CervinoRR23,KrishnagopalR23} further demonstrate the transferability of the gradients of spectral-based GNNs across graphs under similar conditions. Furthermore, using the graphop operator~\cite{backhausz2022action}, \citet{LeJ23} extend the transferability analysis of GNNs to both dense and sparse graphs. Besides, assuming that the $k$-hop ego-graphs are independent and identically drawn, \citet{EGI} derive the transferability of a well-designed GNN based on the differences of $k$-hop ego-graph Laplacian across graphs.

Spatial GNNs generally follow a recursive message-passing scheme~\cite{MPNN}, where each node updates its feature vector by aggregating the message from its local neighborhood. As discussed in~\cite{GRADE,StruRW}, the marginal distribution shifts $\mathrm{P}(X, G)$ between source and target graph domains can be induced by graph structure and individual node attributes (see Fig.~\ref{fig:graph_transfer}(a)-(c)). Notably, three frameworks have been developed to enhance the transferability of spatial GNNs: invariant node representation~\cite{GRADE,YouCWS23,EvoluNet}, structure reweighting~\cite{StruRW,Pair-Align}, and graph Gaussian process~\cite{GraphGP}. 
\begin{enumerate}[(1)]
    \item {\bf\em Invariant Node Representation:} Inspired by the domain adaptation theory~\cite{redko2019advances}, it is theoretically shown~\cite{GRADE,YouCWS23} that the target error can be bounded in terms of the source error and the graph domain discrepancy. The crucial idea of invariant node representation learning is to explicitly minimize the graph domain discrepancy in a latent feature space, thereby enhancing the transferability of spatial GNNs. For example, AdaGCN~\cite{AdaGCN} and UDA-GCN~\cite{UDAGCN} leverage a domain discriminator to learn the domain-invariant node representation learned by the output layer of GNNs. Inspired by the connection between spatial GNNs and Weisfeiler-Lehman graph kernels~\cite{weisfeiler1968reduction,WLKernels}, GRADE~\cite{GRADE} is proposed based on a graph subtree discrepancy measuring the subtree representation induced distribution shifts across graphs. More recently, SpecReg~\cite{YouCWS23} and A2GNN~\cite{A2GNN} further discuss the impact of spectral regularization and asymmetric model architectures on the transferability of GNNs, respectively.
    
    \item {\bf\em Structure Reweighting:} It is noticed~\cite{StruRW} that invariant node representation might lead to sub-optimal solutions under conditional structure shifts. To solve this problem, StruRW~\cite{StruRW} and Pair-Align~\cite{Pair-Align} are proposed to reweigh the edges of the source graph based on the label-oriented node connections of source and target graphs.

    \item {\bf\em Graph Gaussian Process:} Spatial GNNs are equivalent to graph Gaussian processes in the limit as the width of graph neural layers approaches infinity~\cite{NiuA023}. Based on this observation, GraphGP~\cite{GraphGP} is derived from a graph structure-aware neural network in the limit on the layer width, in order to characterize the relationships between nodes across different graph domains. The generalization analysis of GraphGP further reveals the positive correlation between knowledge transferability and graph domain similarity.

\end{enumerate}

\subsubsection{Transferability on Textual Data}
Transfer learning has been widely studied in various natural language processing (NLP) tasks, e.g., text classification~\cite{RuderH18}, question answering~\cite{WieseWN17}, neural machine translation~\cite{zhao2020learning}, etc. A key challenge in understanding textual transferability is the non-IID nature of words/tokens, as they might co-occur within sequences or documents. Thus, recent theoretical analyses of textual transferability often consider an alternative assumption~\cite{lotfi2024nonvacuous}, i.e., sequences or documents are independently drawn from the same distribution. This assumption enables theoretically deriving the transferability and generalization of transfer learning in sequence-level and document-level NLP tasks. Take multilingual machine translation as an example, the goal is to train a single neural machine translation model to translate between multiple source and target languages~\cite{ZophK16}. To achieve this, language-invariant representation learning has been introduced to align the sentence distributions of different languages within a shared latent space~\cite{arivazhagan2019missing}. Nevertheless, \citet{zhao2020learning} theoretically analyze the fundamental limits of language-invariant representation learning by deriving a lower bound (w.r.t. marginal sentence distributions from different languages) on the translation error in the many-to-many language translation setting.

More recently, large language models (LLMs) have revolutionized the field of NLP~\cite{GPT4,T5,GPT3,Llama}. The transferability of LLMs has been studied, as fine-tuning LLMs on downstream tasks has become the {\em de facto} learning paradigm. However, it is computationally expensive and resource-intensive to fine-tune the entire LLM model weights with billions of parameters via gradient-based optimization~\cite{BERT}. To solve this problem, parameter-efficient fine-tuning (PEFT) has been investigated from the perspectives of model tuning~\cite{BitFit,LoRA,Adapters} and prompt tuning~\cite{AutoPrompt,Prefix-Tuning,PromptTuning}. The goal is to adapt LLMs to various downstream tasks by adjusting as few parameters as possible. 

Model tuning based approaches explore model architectures or parameters of LLMs for parameter-efficient fine-tuning~\cite{AghajanyanGZ20}. A variety of parameter-efficient model tuning frameworks have been proposed, including adapters~\cite{Adapters}, low-rank decomposition~\cite{LoRA,Compacter}, and selective masking~\cite{BitFit,DiffPruning}. The key ideas behind these frameworks are illustrated in Fig.~\ref{fig:peft}. In general, these approaches aim to update only a few parameters by inserting new trainable modules, adding low-rank parameter matrices, or modifying specific parameters (e.g., bias terms).

\begin{enumerate}[(1)]
    \item {\bf\em Adapters:}
    Adapter-based approaches add new learnable modules with a small number of parameters to LLMs, e.g., maximizing the likelihood $p(y|x; \theta_{\mathrm{adapter}} \circ \theta_{\mathrm{LLM}})$ with added modules $\theta_{\mathrm{adapter}}$. Initially, inspired by visual adapter modules~\cite{RebuffiBV17}, \citet{Adapters} study the adapter-based fine-tuning mechanism in NLP tasks. This method inserts two adapters sequentially within each Transformer block~\cite{Attention}: one following the self-attention layer and another after the feed-forward layer. Nevertheless, follow-up research~\cite{AdapterFusion,BapnaF19} demonstrates that inserting a single adapter after the feed-forward layer can achieve competitive performance while adding fewer parameters. Furthermore, by highlighting the connections between (model-based) adapters and (prompt-based) Prefix Tuning~\cite{Prefix-Tuning}, \citet{HeZMBN22} introduce a family of parallel adapters that directly condition the adapters at different Transformer layers on the input text. On top of adapters, AdapterDrop~\cite{AdapterDrop} and CoDA~\cite{ConditionalAdapters} improve both fine-tuning and inference efficiency by removing adapters from lower Transformer layers and querying only a small subset of input tokens against the pre-trained LLMs, respectively.

\begin{figure}[!t]
  \centering
  
  \begin{subfigure}{0.47\textwidth}
    \centering
    \includegraphics[width=\textwidth]{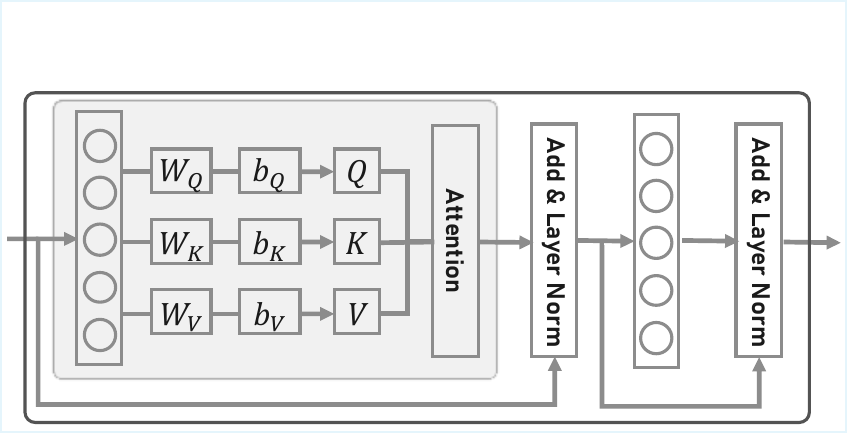}
    \caption{Pre-trained model with Transformer layers}
    \label{fig:a}
  \end{subfigure}
  \begin{subfigure}{0.52\textwidth}
    \centering
    \includegraphics[width=\textwidth]{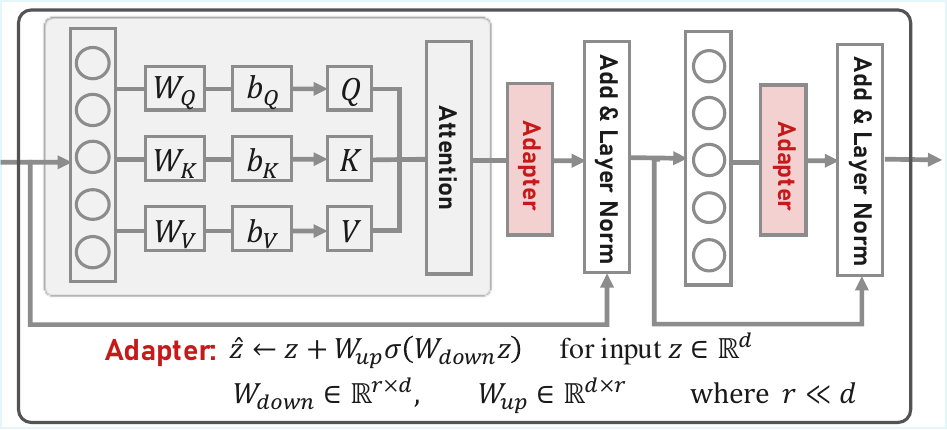}
    \caption{Compositional parameter updates (inserted adapter modules)}
    \label{fig:b}
  \end{subfigure}
  
  \vspace{0.75em} 
  
  \begin{subfigure}{0.47\textwidth}
    \centering
    \includegraphics[width=\textwidth]{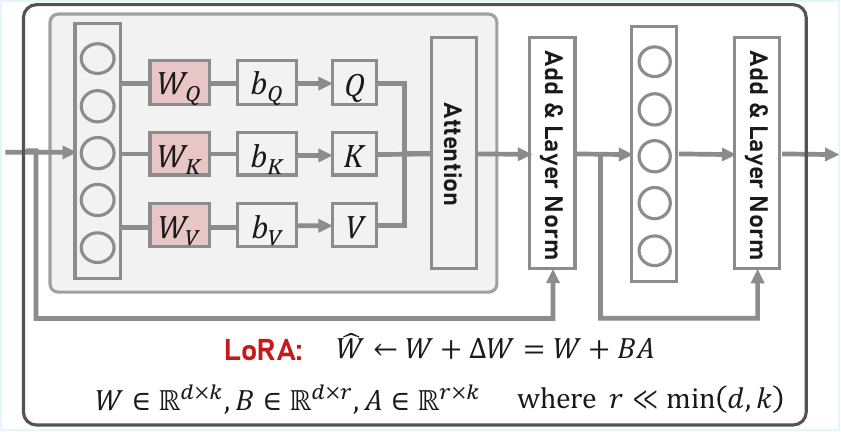}
    \caption{Additive parameter updates (added low-rank $\Delta W$)}
    \label{fig:c}
  \end{subfigure}
  \begin{subfigure}{0.52\textwidth}
    \centering
    \includegraphics[width=\textwidth]{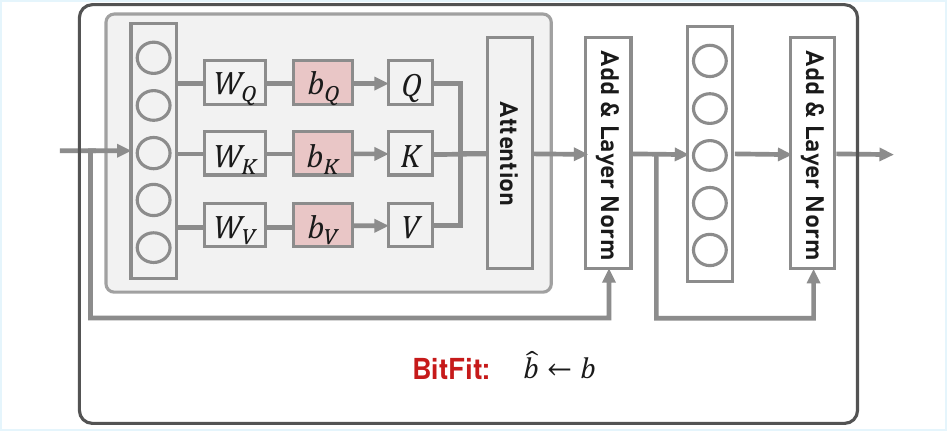}
    \caption{Selective parameter updates (selected bias term only)}
    \label{fig:d}
  \end{subfigure}
  
  \vspace{-2mm}
  \caption{Illustration of parameter-efficient fine-tuning with (b) Adapters~\cite{Adapters} where new modules are inserted, (c) LoRA~\cite{LoRA} where low-rank parameter matrices are added, and (d) BitFit~\cite{BitFit} where only bias terms will be updated.}\label{fig:peft}
    \vspace{-3mm}
\end{figure}


    \item {\bf\em Low-rank Decomposition:}
    Low-rank decomposition injects trainable low-rank decomposition matrices into pre-trained model parameters without changing the model architectures, e.g., maximizing the likelihood $p(y|x; \theta_{\mathrm{LLM}} + \Delta \theta_{\mathrm{LLM}})$ with low-rank $\Delta \theta_{\mathrm{LLM}}$. This line of research is motivated by the phenomenon~\cite{AghajanyanGZ20,LiFLY18} that pre-trained models tend to have a low intrinsic dimension. Here, the intrinsic dimension indicates the lowest dimensional parameter subspace in which satisfactory fine-tuned accuracy on downstream tasks can be achieved. Moreover, by assuming that the parameter change in LLMs during fine-tuning also has a low "intrinsic rank", LoRA~\cite{LoRA} is introduced by optimizing low-rank decomposition matrices of the parameter change. Empirically, LoRA has been further improved in various aspects, including rank selection/optimization~\cite{DyLoRA,AdaLoRA,SoRA}, advanced optimizer~\cite{LoRA+,zhang2024riemannian}, etc. Theoretically, the expressiveness and generalization of LoRA have also been analyzed. Notably, \citet{zeng2024the} prove that under mild conditions regarding the rank of LoRA, it can adapt a pre-trained (or randomly initialized) model to approximate any target model of equal or smaller size. \citet{MalladiWYCA23} find that LoRA fine-tuning is approximately equivalent to full fine-tuning in the Neural Tangent Kernel (NTK) regime~\cite{NTK}, if $r\geq \Theta(\log n^t /\epsilon^2)$ where $r$ is the rank of LoRA and $\epsilon$ is an approximation tolerance. \cite{jang2024lora} show that LoRA fine-tuning has no spurious local minima in the NTK regime, if $r(r+1) > 2Kn^t$ where $K$ is the output dimension, and $n^t$ is the number of training samples in the downstream target task. Furthermore, the generalization bounds of LoRA fine-tuning are theoretically derived in recent works~\cite{jang2024lora,zhu2024asymmetry}.
    
    \item {\bf\em Selective Masking:} The crucial idea of selective masking is to update only a small subset of model parameters during fine-tuning, e.g., maximizing the likelihood $p(y|x; \theta_{\mathrm{LLM}} + \Delta \theta_{\mathrm{LLM}})$ with extremely sparse $\Delta \theta_{\mathrm{LLM}}$. Intuitively, it aims to find a binary mask that automatically selects a small subset of parameters for fine-tuning~\cite{ZhaoLMJS20}. There are three main frameworks for learning this mask. The first one is random masking~\cite{xu2024random,Child-Tuning}, where all the elements of a mask are sampled independently from a Bernoulli distribution. \citet{xu2024random} demonstrate the effectiveness of random masking under a larger-than-expected learning rate, by theoretically building the connection between random masking and flat loss landscape. The second approach involves heuristically-motivated masks, such as bias terms~\cite{BitFit} and cross-attention layers~\cite{Gheini0M21}. The third approach optimizes masks over model parameters from various perspectives, e.g., $L_0$-norm penalty~\cite{DiffPruning}, Fisher information~\cite{FISH-Mask,Child-Tuning,FishDip}, Lottery Ticket Hypothesis~\cite{LT-SFT,PlonerA24}, etc.
\end{enumerate}

Unlike model tuning, prompt tuning keeps the pre-trained LLMs fixed and pretends a sequence of virtual token embeddings (referred to as a trainable prompt) to the input text, e.g., maximizing the likelihood $p(y | [x, z]; \theta_{\mathrm{LLM}})$ for a labeled sample $(x, y)$ where $\theta_{\mathrm{LLM}}$ denotes LLM model parameters and $z$ represents a prompt. Generally, there are three main frameworks for optimizing newly added prompts: soft (continuous) prompt tuning~\cite{Prefix-Tuning,PromptTuning}, hard (discrete) prompt tuning~\cite{AutoPrompt}, and transferable prompt tuning~\cite{SPoT}. 
\begin{enumerate}[(1)]
    \item {\bf\em Soft Prompt Tuning:}
    The key idea of soft prompt tuning is to represent the virtual prompt as continuous-valued token embeddings. It will update only the continuous-valued embeddings of prompts $z$, either in the input embedding layer~\cite{PromptTuning,ResidualPT} or in different layers of the LLMs~\cite{Prefix-Tuning,P-Tuning,WARP,QinE21}. Theoretically, \citet{WeiXM21} studies the connection between prompt tuning and downstream tasks using an underlying latent variable generative model of text. By assuming that input texts are generated by a Hidden Markov Model (HMM), this work models the downstream task as a function of the posterior distribution of the latent variables. It is then shown that prompt tuning enhances the recovery of the ground-truth labeling function in the downstream classification task. Later, \cite{WangC0H23} further show that a carefully constructed pre-trained Transformer can leverage prompt tuning to approximate any sequence-to-sequence function in a Lipschitz function space. They also analyze the restricted expressiveness of prompt tuning compared to model fine-tuning (e.g., LoRA~\cite{LoRA}). 

    \item {\bf\em Hard Prompt Tuning:}
    Although soft prompts can be optimized via gradient-based optimization, \citet{KhashabiLMQ0WHK22} reveal that the learned embeddings of soft prompts do not correspond to any human-readable tokens, thus lacking semantic interpretations. An alternative solution is hard prompt optimization~\cite{AutoPrompt,GrIPS}, which aims to find human-readable prompts from a pre-defined vocabulary. Specifically, AutoPrompt~\cite{AutoPrompt} greedily selects the optimal token for each location in the prompt based on the gradient of the loss w.r.t. the embeddings over labeled training samples. However, the greedy search strategy can result in disfluent and unnatural prompts. To solve this problem, FluentPrompt~\cite{FluentPrompt} and PEZ~\cite{PEZ} utilize projected gradient descent optimization to update all the tokens in the prompt. The crucial idea is to project the learned continuous-valued embeddings to their nearest neighbors in a pre-defined discrete token space, and then use the mapped tokens to calculate the gradient of the loss. Besides, \citet{choi2024hard,RLPrompt} employ gradient-free reinforcement learning based optimization to discover discrete prompts, especially when LLMs are accessible only via APIs (i.e., model gradients and weights are not accessible). 

    \item {\bf\em Transferable Prompt Tuning:}
    Recent studies have also investigated the transferability of prompts~\cite{SPoT}, where soft prompts are first learned from one or more source tasks and then used as the prompt initialization for the target task. This is motivated by the findings~\cite{SPoT,PPT} that a good prompt initialization is crucial for prompt tuning to achieve competitive performance on the target task, compared to model tuning, especially when the model sizes of LLMs are small. Follow-up research~\cite{SuWQCLWWLLL0SZ22} further analyzes the correlation between soft prompt transferability and the overlapping rate of activated neurons. Inspired by multi-task learning~\cite{MisraSGH16}, MPT~\cite{WangPKF0K23} decomposes the soft prompts for source tasks into a shared matrix and low-rank task-specific matrices, and then transfers the shared matrix to the target tasks. Additionally, studies~\cite{SuWQCLWWLLL0SZ22,wu2024zeroshot} have explored the transferability of soft prompts across different language models in zero/few-shot learning settings.
\end{enumerate}


\subsubsection{Transferability on Time Series Data}
A time series is a sequence of observations collected at even intervals of time and ordered chronologically~\cite{chatfield2004timeseries}. Time series has been extensively applied to model non-stationary data in various high-impact domains, such as weather monitoring~\cite{Dish-TS}, financial forecasting~\cite{ZhouZZLH20}, and healthcare~\cite{ragab2023adatime}. The key challenge in time series analysis lies in characterizing the temporal dependencies and non-stationary (i.e., rapidly changing data distribution over time) of time series data. Generally, time series transfer learning involves the following two tasks: time series forecasting~\cite{DAIN,LiuWWL22} and classification~\cite{VRADA}.

\begin{definition}[Time Series Transferability for Forecasting \cite{DAIN}]
    Given a target time-series data set with historical observations, time series transferability for forecasting aims to predict future events by utilizing its own historical observations or relevant knowledge from another source domain under temporal distribution shifts.
\end{definition}

\begin{figure}[!t]
  \centering
  
  \begin{subfigure}{0.51\textwidth}
    \centering
    \includegraphics[width=\textwidth]{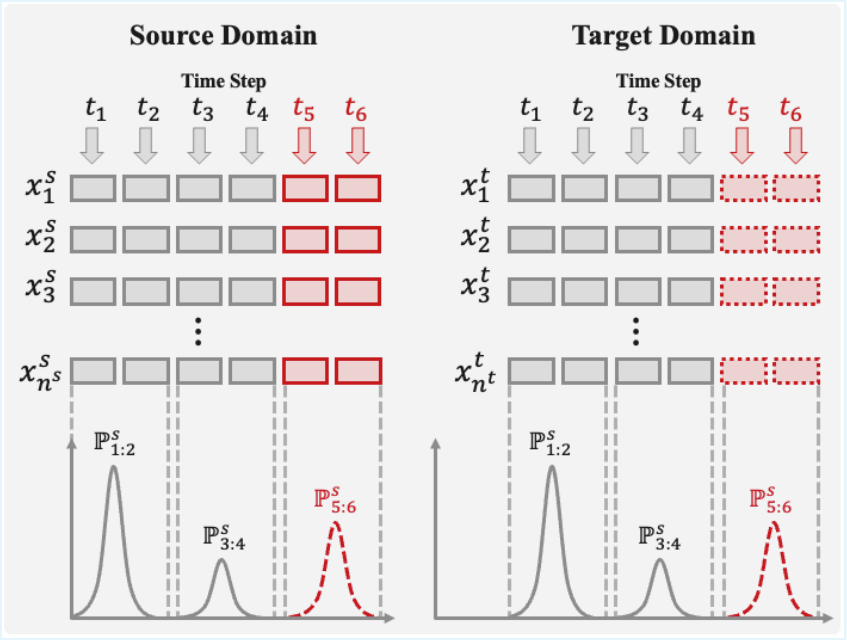}
    \caption{Time series forecasting}
    \label{fig:forecasting}
  \end{subfigure}
  \begin{subfigure}{0.48\textwidth}
    \centering
    \includegraphics[width=\textwidth]{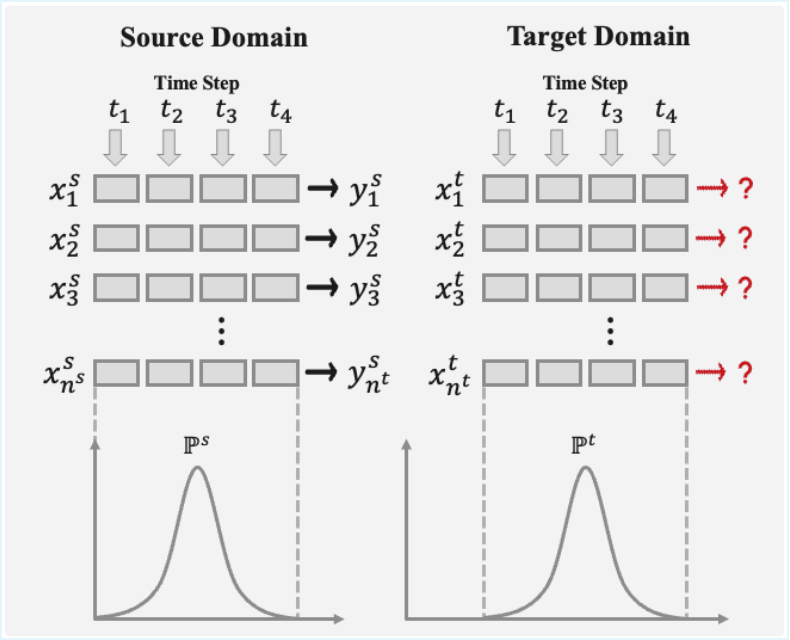}
    \caption{Time series classification}
    \label{fig:classification}
  \end{subfigure}
  
  \vspace{-3mm}
  \caption{Illustration of time series analysis under distribution shifts: (a) time series forecasting, and (b) time series classification}\label{fig:time-series}
    \vspace{-1mm}
\end{figure}

Time series forecasting leverages historical observations to predict future events. As illustrated in Fig.~\ref{fig:time-series}(a), there are two types of distribution shifts within time series forecasting. One is the sample-level temporal distribution shift~\cite{RevIN,LiuWWL22} of non-stationary time series where the data distribution of time series samples changes over time. The other one is the domain-level distribution shifts that occur between source and target time series domains~\cite{DAF}. To address the first type of distribution shifts, AdaRNN~\cite{AdaRNN} characterizes the distribution information by splitting the training sequences into diverse periods with the largest distribution gap, and then dynamically reduces the distribution discrepancy across these identified periods. RevIN~\cite{RevIN} is a symmetrical normalization-and-denormalization method using instance normalization~\cite{ulyanov2016instance}. It first normalizes the input sequences to mitigate distribution shifts among input sequences and then denormalizes the model outputs to restore the statistical information of input sequences. Follow-up approaches such as SAN~\cite{SAN} and Dish-TS~\cite{Dish-TS} build upon RevIN to further address temporal distribution shifts between input and horizon sequences by adaptively learning normalization coefficients for fine-grained temporal slices. More recently, based on Koopman theory~\cite{koopman1931hamiltonian,brunton2022modern}, KNF~\cite{KNF} and Koopa~\cite{Koopa} exploit linear Koopman operators to model the nonlinear dynamics of time series data on the measurement function space. Both methods design a global Koopman operator to learn time-invariant characteristics and a local Koopman operator to capture time-variant dynamics. To address the second type of distribution shifts, DAF~\cite{DAF} uses attention modules to learn complex temporal patterns within time series data and enforces the time-dependent query-key distribution alignment. Particularly, the queries and keys of attention modules are assumed to be domain-invariant, while the values capture domain-specific information for learning domain-dependent time series forecasters.

\begin{definition}[Time Series Transferability for Classification~\cite{VRADA}]
    Given a source domain with labeled time series samples and a target domain with limited or no label information, time series transferability for classification aims to improve the prediction performance of the time series classification model in the target domain by leveraging knowledge from the source domain.
\end{definition}

Time series classification focuses on identifying time series data as a specific category (shown in Fig.~\ref{fig:time-series}(b)). There are two main transfer learning frameworks for time series classification. The first framework involves pre-training a model on a source domain and then fine-tuning it on a target domain. For example, pre-training techniques for time series modeling have been developed using convolutional neural networks~\cite{FawazFWIM18,ConvTimeNet}, recurrent neural networks~\cite{TimeNet}, ResNet~\cite{ZhangZTZ22,SimMTM}, and Transformers~\cite{ZerveasJPBE21}. The second framework is to learn domain-invariant time series representations using adversarial learning~\cite{VRADA,CoDATS,CLUDA,CALDA,CauDiTS} or statistical divergence metrics~\cite{SASA,OttRHBM22,AdvSKM,RAINCOAT} using both source and target data. Specifically, VRADA~\cite{VRADA} captures the domain-invariant temporal latent dependencies of multivariate time-series data using variational recurrent neural networks~\cite{VRNN}, followed by a gradient reversal layer~\cite{DANN}. Similarly, CoDATS~\cite{CoDATS} is developed based on 1D convolutional neural networks and a gradient reversal layer. CLUDA~\cite{CLUDA} and CALDA~\cite{CALDA} further leverage contrastive learning losses to enhance the time series representations. In addition, AdvSKM~\cite{AdvSKM} designs a hybrid spectral kernel network based on maximum mean discrepancy~\cite{MMD} to align source and target time series representations. Assuming that source and target time series domains share the same causal structure, SASA~\cite{SASA} uses long short-term memory (LSTM) networks to learn sparse associative structures from both domains, and then aligns them via maximum mean discrepancy~\cite{MMD}. More recently, CauDiTS~\cite{CauDiTS} furthermore employs an adaptive causal rationale disentanglement to learn domain-invariant causal rationales and domain-specific correlations from variable interrelationships. RAINCOAT~\cite{RAINCOAT} is proposed to address open-world adaptation scenarios, where source and target domains might have domain-specific private classes. It extracts both time features via a 1-dimensional convolutional neural network and frequency features via the discrete Fourier transform, and then aligns the time-frequency features across domains using Sinkhorn divergence~\cite{Cuturi13}.

%% file: content/4_trustworthiness.tex
\section{Knowledge Trustworthiness}\label{sec:trustworthiness}
\begin{figure}
    \centering
    \includegraphics[width=\linewidth]{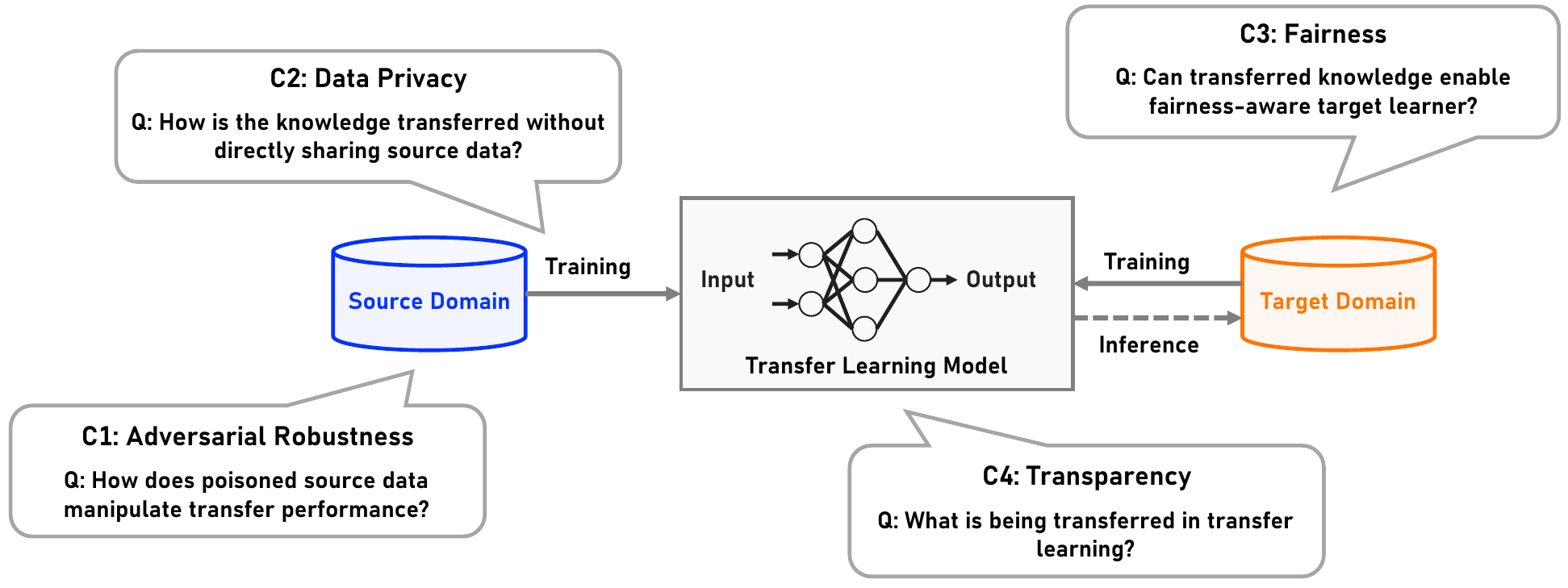}
    \vspace{-6mm}
    \caption{Illustration of trustworthiness concerns in the knowledge transfer process}
    \label{fig:trustworthy_concern}
    \vspace{-3mm}
\end{figure}

In this section, we review the knowledge trustworthiness of transfer learning. Compared to standard trustworthy machine learning over a single domain, this survey discusses whether the source and target users can trust the transferred knowledge, whereas trustworthy machine learning investigates how a user can trust a model trained on private data. As illustrated in Fig.~\ref{fig:trustworthy_concern}, in the context of transfer learning, both the source domain owner and the target domain owner may have trustworthiness concerns about the transfer learning techniques. When considering the owners of the source domain as the ``trustor", do they trust that the transferred knowledge will not leak their data privacy ({\em C1: Privacy})? Conversely, if the "trustor" indicates the owners of the target domain, do they trust that the transferred knowledge is not poisoned ({\em C2: Adversarial Robustness}) or biased ({\em C3: Fairness}), and how well can the transferred knowledge be explained ({\em C4: Transparency})?

\subsection{Privacy}
Privacy protection aims to prevent the unauthorized access or misuse of data that can directly or indirectly reveal sensitive private information, e.g., age, gender, login credential, fingerprint, medical records, etc. In recent years, privacy concerns in understanding the trustworthiness of artificial intelligence (AI) systems have been emphasized in released AI ethics guidelines~\cite{jobin2019global,EuropeanGuidelines} and legal laws (e.g., General Data Protection Regulation (GDPR)~\cite{goodman2017european} and California Consumer Privacy Act (CCPA)~\cite{harding2019understanding}). Maintaining privacy is critical in privacy-sensitive applications, such as patient clinical data analytics~\cite{dayan2021federated} and mobile keyboard prediction~\cite{hard2018federated}. Particularly, privacy protection in transfer learning frameworks focuses on preventing the leakage of private source data during the knowledge transfer process. This concern has inspired privacy-preserving transfer learning frameworks designed to transfer knowledge from a private source domain to a specific target domain while ensuring data privacy. One key principle of these frameworks is that all source data remains stored locally, with only the updated source models/hypotheses being shared securely.

\subsubsection{Hypothesis Transfer}\label{sec:htl}
\begin{figure}[!t]
  \centering
  \begin{subfigure}{0.33\textwidth}
    \centering
    \includegraphics[width=\textwidth]{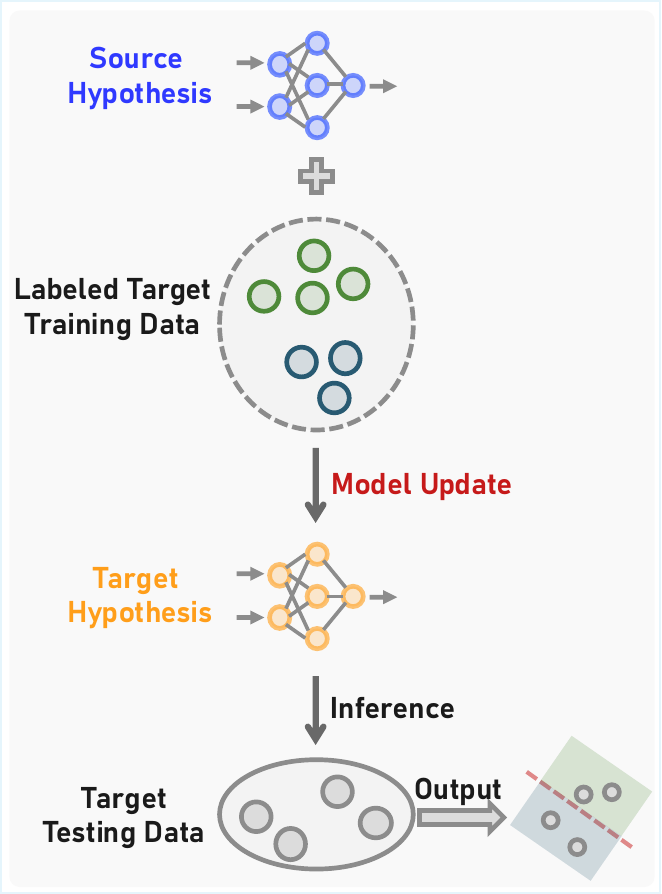}
    \caption{Hypothesis transfer learning}
    \label{fig:hypo_tl}
  \end{subfigure}
  \begin{subfigure}{0.33\textwidth}
    \centering
    \includegraphics[width=\textwidth]{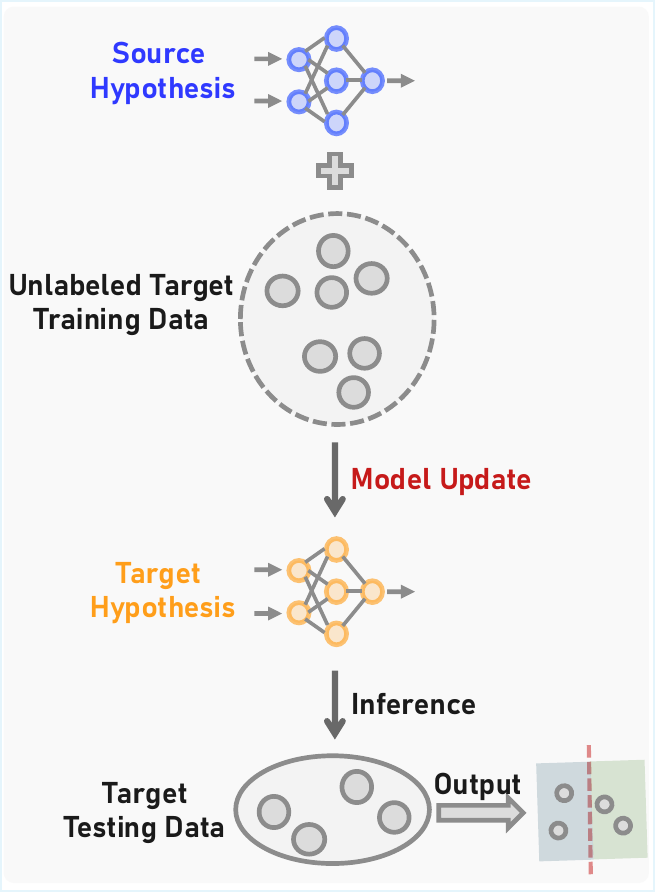}
    \caption{Source-free adaptation}
    \label{fig:sfa}
  \end{subfigure}
  \begin{subfigure}{0.33\textwidth}
    \centering
    \includegraphics[width=\textwidth]{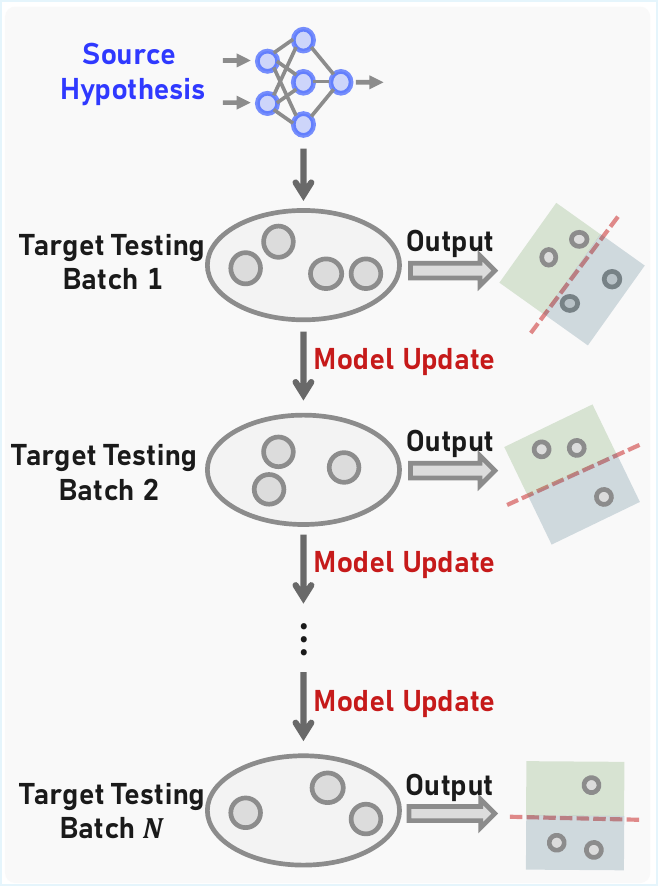}
    \caption{Fully test-time adaptation}
    \label{fig:tta}
  \end{subfigure}

  \vspace{-2mm}
  \caption{Illustration of source hypothesis transfer, including (a) hypothesis transfer learning with labeled target training data, (b) source-free adaptation with unlabeled target training data, and (c) test-time adaptation with only target testing data.}\label{fig:htl}
  \vspace{-2mm}
\end{figure}

Hypothesis transfer involves leveraging the source hypothesis pre-trained from the source data set to solve a learning task on the target domain. It assumes that the target learner has no access to the raw source data or the relatedness between the source and target domains, thereby protecting the data privacy of the source domain. Formally, given a source hypothesis, the problem of hypothesis transfer can be defined as follows.
\begin{definition}[Hypothesis Transfer~\cite{KuzborskijO13}]
    Given a source hypothesis $f_S \in \mathcal{F}_S$ and a target data set $D_T$ with $n_T$ samples, hypothesis transfer algorithms aim to map the source hypothesis $f_S \in \mathcal{F}_S$ and $D_T$ onto a target hypothesis $f_T \in \mathcal{F}_T$:
    \begin{equation}
        A^{\mathrm{htl}}: (\mathcal{X}\times \mathcal{Y})^{n_T} \times \mathcal{F}^s \to \mathcal{F}_T
    \end{equation}
    where $\mathcal{F}^s$ and $\mathcal{F}_T$ denote the hypothesis spaces of source and target domains, respectively.
\end{definition}
As illustrated in Fig.~\ref{fig:htl}, there are three major learning scenarios for hypothesis transferability: (1) hypothesis transfer learning~\cite{KuzborskijO13} with {\em labeled target training data}, (2) source-free adaptation~\cite{SHOT} with {\em unlabeled target training data}, and (3) test-time adaptation~\cite{Tent} with {\em only target testing data}.

\begin{enumerate}[(1)]
    \item {\bf\em Hypothesis Transfer Learning:} The goal of hypothesis transfer learning is to optimize the learning function on the target domain using the basis of hypotheses from the source domain~\cite{KuzborskijO13}. It assumes that both the source hypothesis and a few labeled target samples are accessible during the training of the target model. Earlier works~\cite{fei2006one,li2007bayesian} utilized the source hypothesis as prior knowledge to guide the target learner in Bayesian learning frameworks. \citet{KuzborskijO13} theoretically analyze the generalization error (instantiated with the leave-one-out error) of regularized empirical risk minimization algorithms for hypothesis transfer learning. It is shown that the generalization error is positively correlated with a quantity $\mathbb{E}_{(x,y)\sim \mathbf{P}_T}[\ell(f_S(x), y)]$ measuring how the source hypothesis performs on the target domain, and hypothesis transfer learning enjoys faster convergence rates of generalization errors when a good source hypothesis is provided. Later, \citet{kuzborskij2017fast} extend generalization error bounds of regularized empirical risk minimization with (i) any non-negative smooth loss function, (ii) any strongly convex regularizer, and (iii) a combination of multiple source hypotheses. They further highlight the impact of the quantity $\mathbb{E}_{(x,y)\sim \mathrm{P}_T}[\ell(f_S(x), y)]$ on the transfer performance and propose a principled approach to optimizing the combination of source hypotheses. Instead, \citet{DuKSP17} introduce a notion of transformation function to characterize the relatedness between the source and the target domains. Using this transformation function, they establish excess risk bounds for Kernel Smoothing and Kernel Ridge Regression. \citet{minami2023transfer} further theoretically derive the optimal form of transformation functions under the squared loss scenario. \citet{AghbalouS23} analyze hypothesis transfer learning through regularized empirical risk minimization under reproducing kernel Hilbert space (RKHS) with surrogate classification losses (e.g., exponential loss, logistical loss, softplus loss, mean squared error, and squared hinge) in the context of binary classification. They establish the generalization error bounds and excess risk bounds based on hypothesis stability and pointwise hypothesis stability, highlighting the connections between surrogate classification losses and the quality of the source hypothesis. In addition, \citet{TOHAN,DongLCLG0S023} study a more challenging few-shot hypothesis adaptation problem, where only a few labeled target samples (e.g., one sample per class) are available. Motivated by the learnability of semi-supervised learning, they propose generating highly-compatible unlabeled data to improve the training of the target learner.
    
    It is noteworthy that gradient-based fine-tuning has become the predominant hypothesis transfer approach in the era of large foundation models~\cite{YosinskiCBL14,GoukHP21}. Given a source hypothesis pre-trained on a source domain with adequate labeled samples, gradient-based fine-tuning aims to update the hypothesis through gradient descent optimization using a small amount of labeled target samples. The generalization performance of fine-tuning has been theoretically studied recently~\cite{ShachafBG21,JuLZ22}. \citet{ShachafBG21} show that the generalization error of fine-tuning under certain architectures (e.g., deep linear networks~\cite{JiT19}, shallow ReLU networks~\cite{AroraDHLW19}) can be affected by the difference between optimal (normalized) source and target hypotheses, the covariance structure of the target data, and the depth of the network. Furthermore, recent works~\cite{GoukHP21,LiZ21,JuLZ22} show the generalization error of fine-tuning techniques in terms of the distance between fine-tuned and initialized model parameters.

    \item {\bf\em Source-free Adaptation:} In contrast to hypothesis transfer learning, source-free adaptation~\cite{KunduVVB20,SHOT} enables hypothesis transfer from the source to the target domains using only unlabeled target data. To solve this problem, SHOT~\cite{SHOT} is proposed to update the feature extractor of the pre-trained source model for the target domain while keeping the source classifier fixed. It maximizes the mutual information between intermediate feature representations and the output of the classifier, and also minimizes the prediction error using self-supervised pseudo labels. The follow-up works have developed source-free adaptation frameworks from various perspectives: clustering~\cite{yang2022attracting}, pseudo-labeling~\cite{boudiaf2023search,lee2022confidence}, data augmentation~\cite{KunduKBMKJR22,hwang2024sfda}, etc. As discussed in \cite{mitsuzumi2024understanding}, most existing source-free adaptation approaches focus on understanding the discriminability-diversity trade-off:  the former improves the discriminability of unlabeled target samples in the latent feature space while the latter ensures prediction diversity for all classes. In particular, \citet{mitsuzumi2024understanding} establish a theoretical connection between source-free adaptation and self-training~\cite{wei2021theoretical} in terms of discriminability and diversity losses. This connection enables improved training of source-free adaptation by incorporating an auto-adjusting diversity constraint and teacher-student augmentation learning. In contrast, \citet{KunduKBMKJR22,han2023discriminability} study the discriminability-transferability trade-off in the context of source-free adaptation. 
    Theoretically, \citet{shen2023balancing} derive an information-theoretic generalization error bound for multi-source-free adaptation based on a bias-variance trade-off. Here, bias is triggered by the label and feature misalignments across domains, and variance is triggered by the number of pseudo-labeled target samples. \citet{yi2023when} establish the connections between source-free adaptation and learning with noisy labels, given the findings that the pseudo-labels of target samples generated by the source model can be noisy due to domain shift. They theoretically justify the existence of the early-time training phenomenon (ETP) in source-free adaptation scenarios and propose using early learning regularization~\cite{liu2020early} to prevent the model from memorizing label noise during training. Empirically, in addition to standard vision tasks, \citet{boudiaf2023search} reevaluate existing source-free adaptation methods in a more challenging set of naturally occurring distribution shifts in bioacoustics. Their findings indicate that these existing methods often lack generalizability and perform worse than no adaptation in some cases. This highlights the necessity of evaluating source-free adaptation methods across a range of tasks, data modalities, and degrees of distribution shifts.

    \item {\bf\em Test-time Adaptation:} Fully test-time adaptation~\cite{Tent} aims to adapt the source hypothesis to the target testing data, where data batches arrive sequentially and each batch can only be observed once. To this end, Test-Time Training (TTT)~\cite{sun2020test} and its modified version (TTT++)~\cite{TTT++} incrementally update the feature extractor by minimizing the auxiliary task loss. Notably, this approach requires optimizing both this auxiliary task loss and the standard supervised loss. In contrast, without changing the training phase, Tent~\cite{Tent} is proposed to minimize the entropy of model predictions by only updating the normalization statistics and channel-wise affine transformations in an online manner. The follow-up works further improve this framework from two perspectives. One is to enhance the stability and robustness of test-time adaptation by minimizing the entropy of the average prediction across different augmentations~\cite{zhang2022memo} or by minimizing sharpness-aware entropy~\cite{niu2023towards,gong2023sotta}. The other addresses catastrophic forgetting by regularizing the updated parameters~\cite{EATA,wang2022continual} or adaptive resetting the model parameters~\cite{niloy2024effective}. Furthermore, \citet{GoyalSRK22} analyze test-time adaptation through the lens of convex conjugate loss functions and proposes a principled self-training approach based on conjugate pseudo labels for test-time adaptation. Later, \citet{wang2023towards} theoretically justify the advantages of conjugate labels over hard labels in test-time adaptation by showing the performance gap between gradient descent with conjugate labels and gradient descent with hard labels in a binary classification problem. Empirically, \citet{zhao2023pitfalls} identify the commonly seen pitfalls when evaluating test-time adaptation algorithms, including sensitive hyperparameter selection, inconsistent source hypothesis, and insufficient consideration of various types of distribution shifts. \citet{bao2023adaptive} further demonstrate that the modules (e.g., batch normalization layers~\cite{Tent}, feature extractor layers~\cite{sun2020test}, classifier layers~\cite{iwasawa2021test}) selected for test-time adaptation are strongly correlated with the types of distribution shifts. 
    
\end{enumerate}

\subsubsection{Federated Transfer}
\begin{figure}
    \centering
    \includegraphics[width=\linewidth]{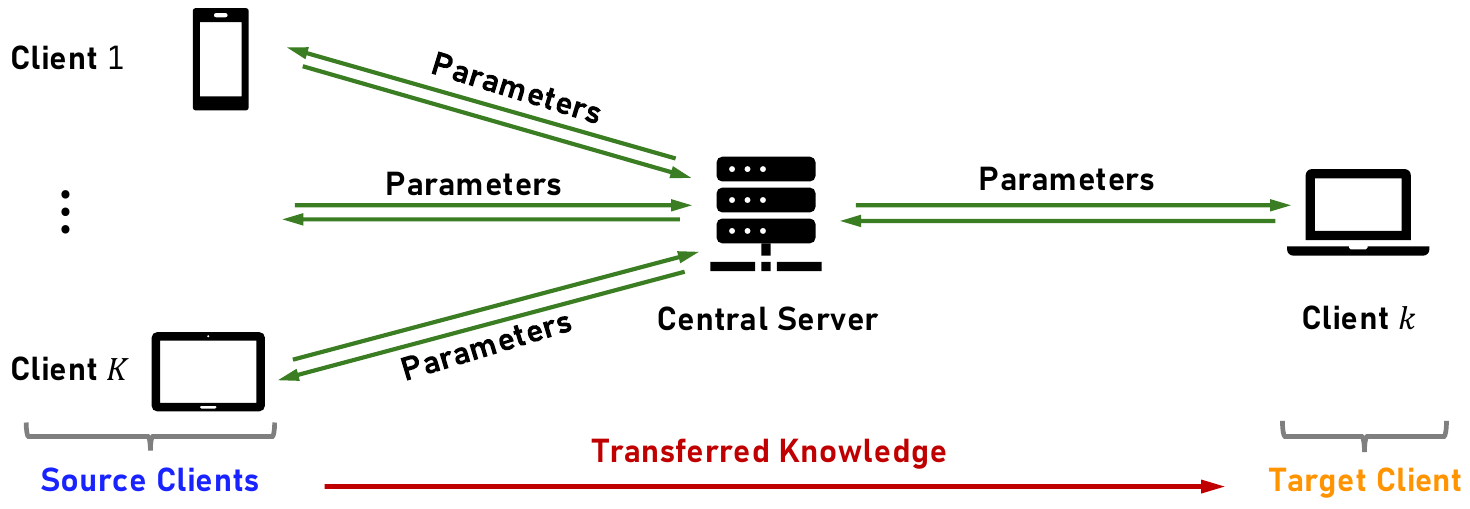}
    \vspace{-7mm}
    \caption{Illustration of personalized federated learning in two scenarios: (1) Generalization performance is evaluated across all clients, where each client (e.g., client $k$ ($k=1,2,\cdots,$)) is considered as a target client and others as source clients for knowledge transfer. (2) Generalization performance is improved only on a specific target client (i.e., only client $k$).}
    \label{fig:fl}
    \vspace{-1mm}
\end{figure}

In contrast to the unidirectional hypothesis transfer discussed in the previous subsection, federated transfer emphasizes bidirectional knowledge sharing that allows source and target domains to communicate and exchange information while maintaining privacy protection~\cite{FedAvg}. This is largely inspired by recent personalized federated learning frameworks~\cite{KairouzMABBBBCC21,liu2020secure} which allow private clients to collaborate in training personalized models under the coordination of a central server. As illustrated in \cite{FedAvg,KairouzMABBBBCC21}, during each communication round, private clients upload their model updates to the central server, which then securely aggregates these updates and broadcasts the updated model back to each client. In this process, each client exclusively owns their data, which will not be shared with the central server or with other clients. From the perspective of knowledge transferability, the intuition behind personalized federated learning is to transfer knowledge across private clients in a privacy-preserving manner~\cite{FedHealth,FEDORA,liu2020secure}. In other words, each private (target) client participates in federated collaboration to receive knowledge from other (source) clients, and its uploaded parameters can be used as indicators to select only the most relevant source knowledge (e.g., related clients with similar parameters within a coalition~\cite{FedCollab,donahue2021model}) under distribution shifts across clients. Formally, given a set of private clients, each with access to a private training set, the problem of federated transfer can be defined as follows.
\begin{definition}[Federated Transfer]
    Given a central server and $K$ private clients each with training samples $D_k$ ($k=1,\cdots,K$), federated transfer aims to learn a personalized model $f_k \in \mathcal{F}_k$ on the $k^{\mathrm{th}}$ client ($k=1,\cdots,K$) by leveraging useful knowledge from the other clients $\{ k' | k'\neq k \}$.
    \begin{align}
        A^{\mathrm{fl}}: (\mathcal{X}\times \mathcal{Y})^{n_k} \times \left( \mathcal{F}_1 \times \cdots \times \mathcal{F}_{k-1} \times \mathcal{F}_{k+1} \times \cdots \times \mathcal{F}_K \right) \to \mathcal{F}_k
    \end{align}
    where $\mathcal{F}_k$ denotes the hypothesis space of the $k^{\mathrm{th}}$ client. The hypotheses from the other clients are often aggregated at the central server and then transferred to the $k^{\mathrm{th}}$ client, i.e., $\left( \mathcal{F}_1 \times \cdots \times \mathcal{F}_{k-1} \times \mathcal{F}_{k+1} \times \cdots \times \mathcal{F}_K \right) \to \mathcal{F}_k$.
\end{definition}
Note that $\mathcal{F}_1 = \cdots = \mathcal{F}_{K}$ implies that all clients will share the same hypothesis space. It can be seen from Fig.~\ref{fig:fl} that there are two different scenarios: one where the generalization performance of all clients is important, and another where only the generalization performance of a target client matters.

To address the first scenario, various personalized federated learning frameworks have been proposed, including parameter decoupling~\cite{FedRep}, model interpolation~\cite{Ditto,pFedMe}, clustering~\cite{CFL,GhoshCYR20}, multi-task learning~\cite{SmithCST17}, meta-learning~\cite{Per-FedAvg}, knowledge distillation~\cite{zhang2021parameterized,zhu2021data}, Bayesian learning~\cite{achituve2021personalized,zhang2022personalized}, etc. 
Despite the impressive performance of these personalized federated learning frameworks across various applications, it is shown~\cite{FEDORA,FedCollab} that some clients might suffer from negative transfer in the context of personalized federated learning. It implies that their performance can be worse compared to when they train a model solely on their local data without communicating information with other clients. To mitigate negative transfer issues, INCFL~\cite{cho2022to} is proposed to maximize the incentivized client participation by dynamically adjusting the aggregation weight assigned to each client. FedCollab~\cite{FedCollab} optimizes the collaboration structure by clustering clients into non-overlapping coalitions based on their distribution distances and data quantities. Similarly, FEDORA~\cite{FEDORA} adaptively aggregates relevant source knowledge by considering distribution similarities among clients and regularizes local models when the received knowledge has a positive impact on the generalization performance. DisentAFL~\cite{chen2024disentanglement} uses a two-stage knowledge disentanglement and gating mechanism to enhance positive transfer under complex client heterogeneity, e.g., modality heterogeneity, task heterogeneity, and domain heterogeneity among clients.

To address the second scenario, federated domain adaptation~\cite{PengHZS20,Fan0DH23,jiang2024principled} has been studied to transfer knowledge from multiple source clients with sufficient labeled samples to a target client with limited or no labeled samples. Unlike standard personalized federated learning, it focuses only on the generalization performance of the target clients. Specifically, inspired by domain adaptation theory~\cite{A_distance}, \citet{PengHZS20} derive a weighted error bound for federated domain adaptation. Based on this, the FADA algorithm is proposed to disentangle domain-invariant and domain-specific features for each client and then align the domain-invariant features between source and target clients. Similarly, \citet{feng2021kd3a} leverage knowledge distillation and BatchNorm Maximum Mean Discrepancy (MMD) to address the distribution gaps between source and target clients. More recently, \citet{jiang2024principled} theoretically analyze the connections between the generalization performance and aggregation rules of federated domain adaptation. This finding also results in an auto-weighting scheme for optimal combinations of the source and target gradients. In addition to federated domain adaptation, federated domain generalization aims to train models using source clients and then apply these models to previously unseen target clients~\cite{nguyen2022fedsr,yuan2022what}. Notably, \citet{bai2024benchmarking} propose a federated domain generalization benchmark, highlighting the necessity of evaluation scenarios that involve a large number of private clients, high client heterogeneity, and more realistic data sets.

\subsection{Adversarial Robustness}
It has been observed~\cite{SzegedyZSBEGF13,GoodfellowSS14} that modern machine learning models can be easily fooled by adversarial examples that are perceptibly indistinguishable with respect to clean inputs. This survey focuses on exploring the adversarial robustness of knowledge transfer models under assumptions where distribution shifts occur across domains.

\subsubsection{Attacks}\label{sec:attacks}
Recent efforts have been devoted to understanding the adversarial vulnerability of deep transfer learning techniques~\cite{WangYVZZ18,ZhangSLB020,Rezaei2020A}. In the context of transfer learning, evasion attacks aim to generate adversarial examples to fool the learned transfer learning models on the target domain. Initially, by minimizing the feature representation dissimilarity between adversarial and clean target samples from different classes using only the pre-trained model, \citet{WangYVZZ18} demonstrate the vulnerability of fine-tuned models in the transfer learning framework. Based on the observations that the neurons of the activation vector within the pre-trained model correlate with target classes, \citet{Rezaei2020A} design a simple brute-force attacking mechanism. This approach crafts input data to trigger those neurons individually, thereby exploring which one is highly associated with each target class.

\begin{figure}[!t]
  \centering
  \begin{subfigure}{0.59\textwidth}
    \centering
    \includegraphics[width=\textwidth]{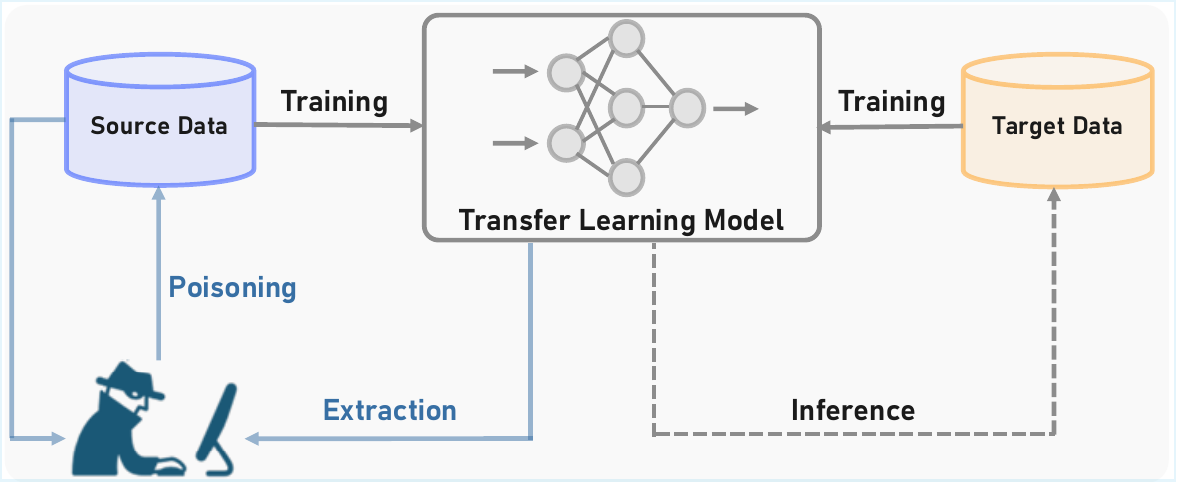}
    \caption{Poisoning attacks on transfer learning}
    \label{fig:tlattack_a}
  \end{subfigure}
  \begin{subfigure}{0.2\textwidth}
    \centering
    \includegraphics[width=\textwidth]{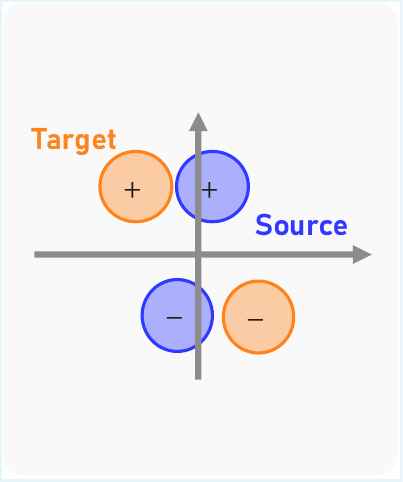}
    \caption{Before attack}
    \label{fig:tlattack_b}
  \end{subfigure}
  \begin{subfigure}{0.2\textwidth}
    \centering
    \includegraphics[width=\textwidth]{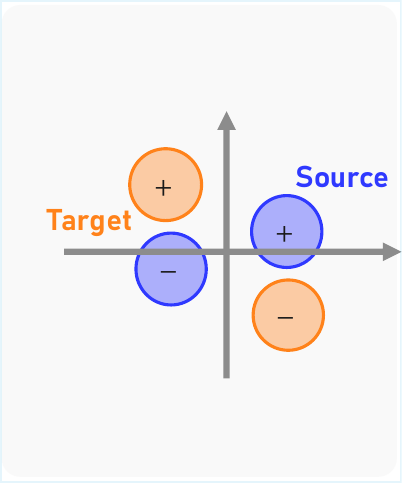}
    \caption{After attack}
    \label{fig:tlattack_c}
  \end{subfigure}

  \vspace{-2mm}
  \caption{Illustration of poisoning attacks on transfer learning. (a) By injecting adversarial noise into the source data, the adversary can control the prediction behavior of transfer learning models on the target domain. One intuitive explanation~\cite{I2Attack,MehraKCH21} is that (b) initially the source and target distributions can be correctly aligned, but (c) they become misaligned after applying the attack.}\label{fig:tlattack}
  \vspace{-2mm}
\end{figure}


In contrast, as shown in Fig.~\ref{fig:tlattack}, poisoning attacks allow crafting source samples to control the prediction behavior of transfer learning models on the target domain during model training. Generally, poisoning attacks can occur in two transfer learning scenarios. The first scenario is the joint training of source and target data, assuming that source and target domains have the same labeling space, and the target domain has only unlabeled training samples (i.e., unsupervised domain adaptation~\cite{TL_survey_2010}). It is motivated by the findings~\cite{0002CZG19} that the feature-based marginal distribution matching can result in negative transfer when the target domain has no label information. Notably, I2Attack~\cite{I2Attack} and AdaptAttack~\cite{AdaptAttack} maximize the label-informed joint distribution discrepancy between the raw source domain and the poisoned source domain with the following constraints. (1) {\em Perceptibly Unnoticeable:} All the poisoned input images are natural-looking. (2) {\em Adversarially Indirect:} Only source samples are maliciously manipulated. (3) {\em Algorithmically Invisible:} Neither source classification error nor marginal domain discrepancy between source and target domains increases. These constraints imply that in the context of transfer learning, an adversary could potentially manipulate the source data to gain control over the prediction function on the target domain. Similarly, \citet{MehraKCH21} propose to generate poisoned source samples with clean labels or mislabeled source samples to fool the discrepancy-based adaptation approaches. 

The second scenario is the pre-training and fine-tuning framework, where a model is first pre-trained on the source domain and then fine-tuned on the downstream target domain. In this scenario, backdoor attacks are intended to manipulate pre-trained model weights, thus resulting in malicious prediction behavior of fine-tuned models on the target domain~\cite{JiZJLW18,ShenJ0LCSFYW21}. The intuition behind backdoor attacks is that when the triggers (e.g., keywords) are activated on target samples, the fine-tuned model will predict pre-defined class labels~\cite{BadNets}. Specifically, backdoor attacks in pre-trained models satisfy the following conditions~\cite{RIPPLe,YaoLZZ19}. (1) Only pre-trained model weights are manipulated, and the infection should be done on the target data through transfer learning. (2) With poisoned pre-trained models, fine-tuned models will behave normally on clean target data, but misclassify any sample with the trigger into a specific class. (3) The designed attacks should be unnoticeable from the viewpoint of the target learner, i.e., the attacker does not alter the fine-tuning process and the training data on the target domain. With these conditions in mind, BadNets~\cite{BadNets} directly uses a poisoned data set to adjust the parameters of pre-trained models. RIPPLe~\cite{RIPPLe} poisons the weights of pre-trained models using a bi-level optimization objective over both the poisoning and fine-tuning losses. However, it is noticed~\cite{RIPPLe} that the fine-tuned models can mitigate the impact of backdoor attacks during the fine-tuning process on clean target samples. Thus, \citet{YaoLZZ19,LiSLZMQ21} focus on poisoning only the lower layers of pre-trained models. Furthermore, \citet{BadEdit} reformulate backdoor injection as a lightweight knowledge editing problem and adjusts only a subset of model parameters (e.g., key-value pairs) with a minimal amount of poisoned data. Rather than manipulating pre-trained model weights, recent works explore backdoor attacks on large language models (LLMs) in the fine-tuning phase by inserting triggers into instructions~\cite{xu2024instructions,yan2024backdooring,ShuW0GXG23} or prompts~\cite{jiang2023forcing,ZhaoWLZF23,BadPrompt}. Besides, it is empirically observed~\cite{WanWSK23,bowen2024scaling} that larger LLMs are more susceptible to poisoning attacks than smaller ones. All these backdoor attacks above highlight security and ethical concerns in developing and deploying pre-trained models~\cite{hubinger2024sleeper}.

\subsubsection{Defenses}
In the context of transfer learning, the adversarial robustness of the prediction function on the target domain can be improved in various scenarios. (1) Given an adversarially pre-trained source model, the adversarial robustness of the source model can be transferred to the target domain~\cite{HendrycksLM19,ShafahiSZGSJG20}. (2) Given a standard pre-trained model, the adversarial robustness of the target learner can be enhanced via robust fine-tuning~\cite{AutoLoRa}. (3) Given an attacked pre-trained model, the defense mechanism can be developed to mitigate the negative impact of source knowledge on the target domain during fine-tuning~\cite{ChinZM21,MDP}.

Recent works~\cite{davchev2019empirical,HendrycksLM19,Chen0C0AW20} empirically demonstrate the transferability of adversarial robustness across domains, e.g., the robustness of an adversarially pre-trained source model can be transferred to the target domain. Specifically, based on the Learning without Forgetting (LoF) approach~\cite{li2017learning}, \citet{ShafahiSZGSJG20,VaishnaviER22} use the distillation regularization to preserve the robust feature representations of the source model during fine-tuning. The intuition is that the lower layers of an adversarially pre-trained source model can capture robust features from input samples. Similarly, to enhance the transferability of adversarial robustness across domains, \citet{RFA} utilize knowledge distillation to preserve the feature correlations of the robust source model on the target domain. \citet{CARTL} propose enforcing feature similarity between natural samples and their corresponding adversarial counterparts during pre-training and regularizing the Lipschitz constant of neural networks during fine-tuning. \citet{TWINS} propose a TWINS structure to incorporate the means and variances of batch normalization layers over both pre-training and target data during adversarial fine-tuning. Notably, the studies mentioned above focus on the transferability of empirical adversarial robustness through adversarial training techniques~\cite{MadryMSTV18,GoodfellowSS14}, which minimize the adversarial objective against pre-determined strong attacks. In addition to empirical adversarial robustness, \citet{alhamoud2023generalizability,vaishnavi2024a} further investigate certified/provable adversarial robustness~\cite{CohenRK19,JeongS20} in the context of transfer learning, which seeks to maximize the radius around inputs within which the model output remains consistent. Theoretically, \citet{NernRGS23} show that the transferability of adversarial robustness can be guaranteed if the feature extractor of the pre-trained source model is robust and only the newly added linear predictor is updated during fine-tuning. This analysis is consistent with the empirical observations~\cite{ShafahiSZGSJG20,hua2024initialization} that feature extractors from an adversarially pre-trained source model contribute to robustness transfer across domains, where only the last layer is re-trained on the target data.

The second line of research is robust fine-tuning~\cite{DongLLYZ21,AutoLoRa}, where a standard pre-trained model is fine-tuned using adversarial training~\cite{MadryMSTV18,GoodfellowSS14}. Specifically, RIFT~\cite{DongLLYZ21} maximizes the mutual information between the feature extracted by the adversarially fine-tuned model and the class label plus the feature extracted by the pre-trained model. AutoLoRa~\cite{AutoLoRa} disentangles robust fine-tuning via a low-rank branch to mitigate gradient conflicts between adversarial and natural objectives. It optimizes adversarial objective w.r.t. the standard feature extractor and standard objective w.r.t. the auxiliary LoRA branch. More recently, \citet{wang2024adversarially} examine the adversarial robustness of hypothesis transfer learning~\cite{KuzborskijO13}, which involves transferring knowledge from source domains to a target domain using a set of pre-trained auxiliary hypotheses. They derive generalization error bounds for adversarial robustness in the target domain based on two specific algorithms: adversarial regularized empirical risk minimization and proximal stochastic adversarial training.

The previous two scenarios assume the availability of a clean pre-trained source model for transfer learning. Their goal is to improve the adversarial robustness of fine-tuned models against adversarial perturbations in the target samples during inference. As discussed in Subsection~\ref{sec:attacks}, a more challenging yet realistic scenario occurs when a poisoned source model~\cite{Rezaei2020A,BadPrompt} is deployed for transfer learning. In this scenario, defense mechanisms should handle the negative impact of poisoned source knowledge during fine-tuning. To this end, \citet{ChinZM21} design a defense mechanism to counter the attack proposed by \citet{Rezaei2020A}. The key idea is to reduce the similarity between the pre-trained and the fine-tuned models via noisy feature distillation. More recently, in the context of backdoored large language models (LLMs)~\cite{BadPrompt,ZhaoWLZF23}, \citet{MDP} propose to detect poisoned target samples associated with triggers during inference by leveraging the different masking-sensitivity of poisoned and clean samples. The intuition is that poisoned samples are more sensitive to random masking than clean samples, as fine-tuned LLMs might exhibit significant changes in predictions when the trigger and normal content are masked within poisoned samples. Similarly, \citet{QiCLYLS21,YangLLZS21} detect poisoned samples using the perplexity changes of samples under word deletion or different robustness properties of clean and poisoned samples against triggers, respectively.

\subsubsection{Transferability vs. Robustness}
In addition to highlighting the adversarial vulnerability and robustness of transfer learning frameworks, recent studies have also explored the connection between knowledge transferability and adversarial robustness~\cite{SalmanIEKM20,TerziAMS21}. To be specific, it is empirically demonstrated~\cite{SalmanIEKM20} that adversarially robust models can transfer better (i.e., higher transfer accuracy in the target domain) than their standard-trained counterparts. That is, though robustness may be at odds with accuracy within the same domain~\cite{TsiprasSETM19}, the adversarial robustness achieved in a source domain can improve the transfer accuracy in a related target domain. \citet{UtreraKEKM21} further explain that in image classification tasks, adversarial training in the source domain biases the learned representations towards retaining shapes, thereby improving transferability in the target domain. Theoretically, \citet{TerziAMS21} provide an information-theoretic justification for adversarial training, implying the trade-off between accuracy on the source domain and transferability on a related target domain. More rigorously, \citet{DengZVKZ21} demonstrate that, for a learning function based on a two-layer linear neural network, adversarially robust representation learning over multiple source domains leads to much tighter transfer error bounds on the target domain than standard representation learning. Alternatively, \citet{XuZMSKL22} show that adversarial training regularizes the function class of feature representation learning, thus improving knowledge transferability across domains.

\subsection{Fairness}
Fairness involves eliminating discrimination when training machine learning models~\cite{eshete2021making,castelnovo2022clarification}. In the legal domain, potential discrimination is defined as disparate treatment (triggered by intentionally treating an individual differently) and disparate impact (triggered by negatively affecting members of a protected group)~\cite{pessach2023algorithmic}. Motivated by this definition, different measures of algorithmic fairness have been proposed in machine learning communities, e.g., individual fairness~\cite{DworkHPRZ12}, group fairness~\cite{FeldmanFMSV15}, 
etc. To be specific, individual fairness~\cite{DworkHPRZ12} maintains that similar individuals should be treated similarly. Group fairness~\cite{FeldmanFMSV15,HardtPNS16} ensures statistical parity among groups with sensitive attributes (e.g., race, gender, age). 
In the context of transfer learning, a fundamental concern is whether the fairness of a machine learning model can be transferred across domains under distribution shifts. Following~\cite{schumann2019transfer,chen2022fairness}, the problem of fairness transfer can be formulated as follows.

\begin{definition}[Fairness Transfer~\cite{schumann2019transfer,chen2022fairness}]
    Given a source domain and a target domain, we denote the fairness violation measures as $\Delta^*_S(\cdot)$ for the source domain and $\Delta^*_T(\cdot)$ for the target domain. For any hypothesis $f\in \mathcal{F}$, algorithmic fairness can be transferred if the following condition is satisfied:
    \begin{equation}
        \Delta^*_T(f) \leq \Delta^*_S(f) + \delta
    \end{equation}
    where $\delta$ quantifies the distribution shifts between the source and target domains.
\end{definition}

\subsubsection{Group Fairness}
Generally, group fairness~\cite{FeldmanFMSV15,HardtPNS16,castelnovo2022clarification} requires that different groups are treated equally. There are several commonly used group fairness metrics: demographic parity~\cite{FeldmanFMSV15}, equality of opportunity~\cite{HardtPNS16}, and equalized odds~\cite{HardtPNS16}. Following~\cite{MadrasCPZ18,schumann2019transfer}, we formally define these metrics in a binary classification problem where $\mathcal{Y} = \{0, 1\}$. Assuming there are two groups defined by binary sensitive attributes $A\in \{0, 1\}$, fair machine learning seeks to ensure accurate predictions without bias against any particular group.
\begin{itemize}
    \item {\em Demographic Parity~\cite{FeldmanFMSV15}:} Demographic parity, also known as statistical parity, requires the same positive prediction ratio across groups with different sensitive attributes.
    \begin{equation}
        \mathrm{Pr}(\hat{Y}=1 | A = 0) = \mathrm{Pr}(\hat{Y}=1 | A = 1)
    \end{equation}
    where $\hat{Y}$ denotes the random variable of the predicted class label and $\mathrm{Pr}(\cdot)$ represents the probability. This criterion implies that the decisions made by machine learning models should be independent of any sensitive attributes. However, it may be limited in scenarios where the base rates of the two groups differ, i.e., $\mathrm{Pr}({Y}=1 | A = 0) \neq \mathrm{Pr}({Y}=1 | A = 1)$ where $Y$ is the ground-truth class variable. In such cases, it is unrealistic to expect both model accuracy and demographic parity to be achieved simultaneously. Notably, \citet{ZhaoG19} theoretically characterize the inherent trade-off between statistical parity and prediction accuracy, by providing a lower bound on group-wise prediction error for any fair predictor under demographic parity.

    \item {\em Equality of Opportunity~\cite{HardtPNS16}:} A machine learning model is considered fair under equality of opportunity if the false positive rates across groups are equal.
    \begin{equation}
        \mathrm{Pr}(\hat{Y}=1 | A = 0, Y=0) = \mathrm{Pr}(\hat{Y}=1 | A = 1, Y=0)
    \end{equation}
    In contrast to demographic parity, equality of opportunity considers the ground-truth class variable $Y$. It enables the base rates for the two groups to be different. Similarly, a symmetric definition can be formulated using the false negative rates, i.e., $\mathrm{Pr}(\hat{Y}=0 | A = 0, Y=1) = \mathrm{Pr}(\hat{Y}=0 | A = 1, Y=1)$.

    \item {\em Equalized Odds~\cite{HardtPNS16}:} A machine learning model is considered fair under equalized odds if both the false positive rates and false negative rates across groups are equal.
    \begin{equation}
    \begin{aligned}
        \mathrm{Pr}(\hat{Y}=1 | A = 0, Y=0) &= \mathrm{Pr}(\hat{Y}=1 | A = 1, Y=0) \\
        \mathrm{Pr}(\hat{Y}=0 | A = 0, Y=1) &= \mathrm{Pr}(\hat{Y}=0 | A = 1, Y=1)
    \end{aligned}
    \end{equation}
    
\end{itemize}

\begin{figure}[!t]
  \centering
  \begin{subfigure}{0.31\textwidth}
    \centering
    \includegraphics[width=\textwidth]{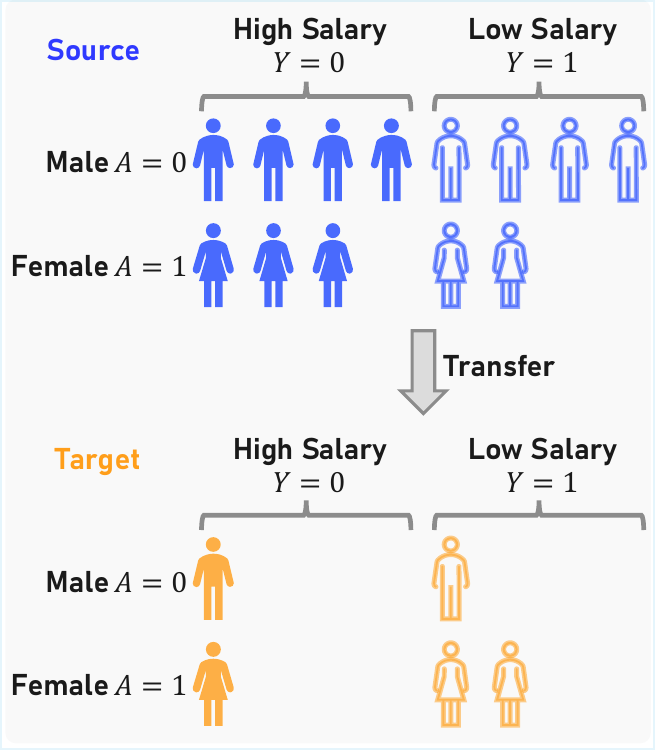}
    \caption{Fairness transfer with $(X_T, A_T, Y_T)$}
    \label{fig:group_fairness_a}
  \end{subfigure}
  \begin{subfigure}{0.31\textwidth}
    \centering
    \includegraphics[width=\textwidth]{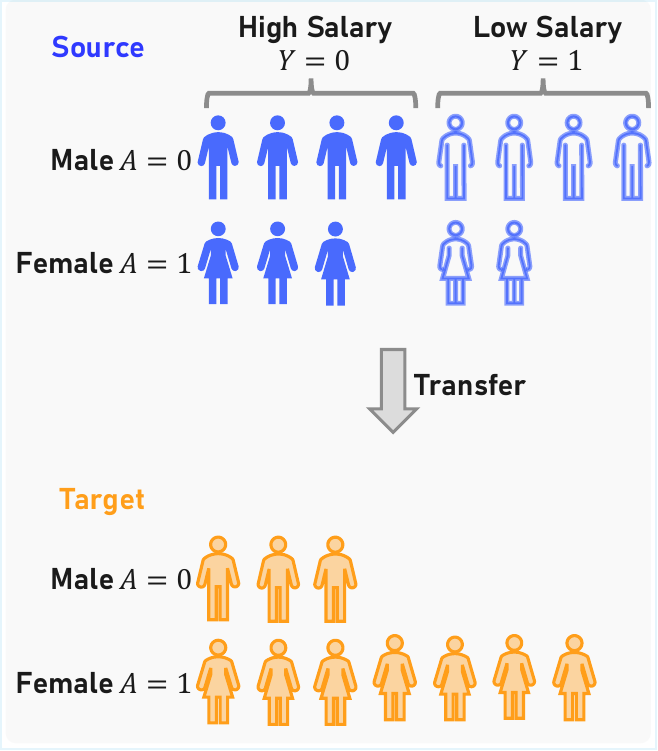}
    \caption{Fairness transfer with $(X_T, A_T)$}
    \label{fig:group_fairness_b}
  \end{subfigure}
  \begin{subfigure}{0.31\textwidth}
    \centering
    \includegraphics[width=\textwidth]{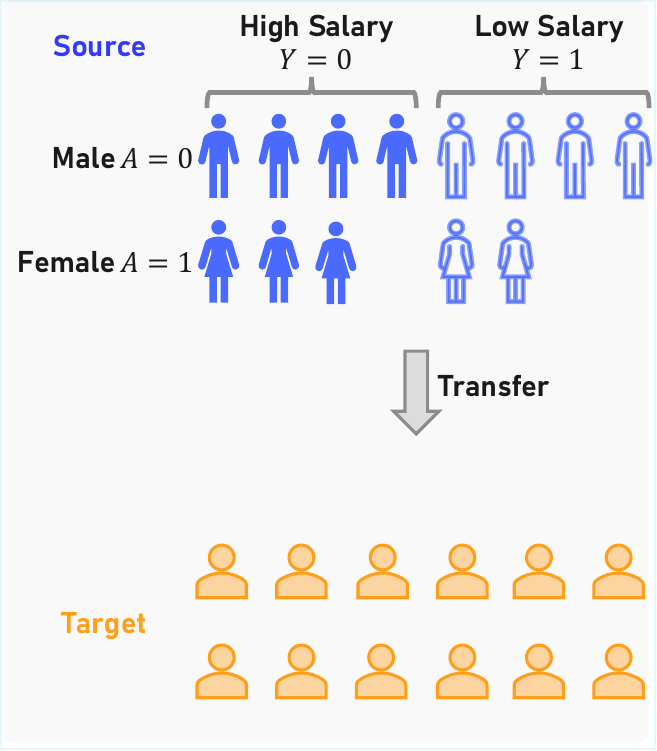}
    \caption{Fairness transfer with $X_T$}
    \label{fig:group_fairness_c}
  \end{subfigure}

  \vspace{-3mm}
  \caption{Transferability of group fairness across domains. (a) The target domain has labeled samples with sensitive attributes. (b) The target domain has unlabeled samples with sensitive attributes. (c) The target domain has only unlabeled samples without sensitive attributes.}\label{fig:group_fairness}
  \vspace{-2mm}
\end{figure}

Fair transfer learning that integrates the aforementioned group fairness criteria has been studied in recent years~\cite{giguere2022fairness,dutt2024fairtune}. For example, \citet{zemel2013learning,MadrasCPZ18} propose learning fair intermediate representations by encoding the data as accurately as possible while obscuring information about sensitive attributes. They demonstrate the transferability of these fair representations across different tasks. Later, \citet{SchrouffHKASOBR22} empirically investigate the connections between compound distribution shifts (e.g., the co-occurrence of demographic, covariate, and label shifts) and fairness transfer in real-world medical applications via a joint causal framework~\cite{mooij2020joint}. Furthermore, \citet{chen2022fairness} provide a generic Lipshitz upper bound for group fairness when the underlying distribution shifts (e.g., covariate shift or label shift between source and target domains) are constrained. Specifically, most existing works~\cite{coston2019fair,biswas2021ensuring,zhao2024algorithmic} dive into understanding the transferability of fairness, by considering various learning scenarios based on the availability of class labels and sensitive attribute information in the target domain. Fig.~\ref{fig:group_fairness} illustrates three scenarios for group fairness transfer when training data are available in the target domain. Another related scenario is the domain generalization~\cite{pham2023fairness} where no target samples are available during training.

\begin{enumerate}[(1)]
    \item {\em Labeled target samples with sensitive attributes:} Assuming that the target domain contains a few labeled samples with sensitive attributes, \citet{schumann2019transfer} provide the generalization error bounds of group fairness (e.g., equality of opportunity, and equalized odds) in the target domain in terms of fairness-aware distribution discrepancy between source and target domains. \citet{oneto2020learning,oneto2020exploiting} theoretically show the generalization error bound of group fairness (e.g., demographic parity) across domains from the perspective of multi-task learning via low-rank matrix factorization or parameter decoupling. Similarly, \citet{slack2020fairness} propose a fair meta-learning algorithm to transfer the fairness across domains.

    \item {\em Unlabeled target samples with sensitive attributes:} When the target domain has only unlabeled samples with sensitive attributes, \citet{rezaei2021robust} propose minimizing both the expected log-loss and the pseudo-label aware fairness penalty over the worst-case approximation of the target distribution to mitigate covariate shifts across domains and ensure fairness (e.g., demographic parity, equality of opportunity, and equalized odds) in the target domain. \citet{havaldar2024fairness} leverage representation matching across sensitive groups to enforce fairness and sample reweighting to mitigate covariate shifts across domains. Inspired by the theory of self-training~\cite{wei2021theoretical,cai2021theory}, \citet{AnCDH22} theoretically analyze the transferability of group fairness across domains based on the consistency loss of a machine learning model under input transformations. Then they propose a self-training algorithm with fair consistency regularization to improve fairness transfer in the presence of subpopulation shifts. In contrast, \citet{roh2023improving} formalize the notion of correlation shift over labels and sensitive attributes and employs a weighted sampling strategy in data preprocessing to mitigate correlation shifts across domains.

    \item {\em Only unlabeled target samples with missing sensitive attributes:} \citet{coston2019fair} study a more general learning scenario where the target domain is associated with only unlabeled samples with missing sensitive attributes. To improve group fairness (e.g., demographic parity) in the target domain, they develop fairness-guided sample reweighting approaches by enforcing the similarity of group-wise weighting scores across all pairs of groups.

    \item {\em No target samples:} An extreme situation occurs when no target samples are available, commonly referred to as domain generalization or out-of-distribution generalization~\cite{BlanchardLS11,gulrajani2021in}. In this scenario, only source domain data is provided to learn a fair predictor for unseen target domains. To solve this problem, \citet{singh2021fairness} develop a causal inference framework to minimize the worst-case prediction error under group fairness constraints. Similarly, \citet{mandal2020ensuring} focus on optimizing a fair predictor by minimizing the worst-case error across weighted combinations of the training data. Later, \citet{pham2023fairness} derive the theoretical upper bounds on generalization error and unfairness in the target domain in terms of source error/unfairness, the domain discrepancy among source domains, and the domain discrepancy between source and unseen target domains. Motivated by this theoretical analysis, they propose an invariant representation learning algorithm to improve the transfer of fairness and accuracy via density matching.
\end{enumerate}

\subsubsection{Individual Fairness}
Individual fairness requires that similar individuals (in the input space) should receive similar decision outcomes (in the output space)~\cite{DworkHPRZ12,zemel2013learning}. Individuals are similar if their only differences lie in protected attributes or features related to those attributes. Mathematically, \citet{DworkHPRZ12} formalize this notion using $L$-Lipschitz continuity of a function $f: \mathcal{X}\to \mathcal{Y}$. For all $x_1, x_2 \in \mathcal{X}$, the following holds
\begin{equation}
    d_{\mathcal{Y}}\left( f\left(x_1\right), f\left(x_2\right) \right) \leq L \cdot d_{\mathcal{X}}\left( x_1, x_2 \right)
\end{equation}
where $L$ is a constant. Here, $d_{\mathcal{X}}$ and $d_{\mathcal{Y}}$ represent the distance metrics in the input space and output space, respectively. Recently, \citet{MukherjeePYS22} investigate the connections between individual fairness and knowledge transferability in unsupervised domain adaptation/generalization scenarios. They show that (i) enforcing individual fairness (e.g., graph Laplacian regularizer~\cite{KangHMT20}) can theoretically improve the generalization performance of a learning function under the covariate shift assumption, and (ii) invariant representation learning commonly used in existing domain adaptation algorithms~\cite{DANN} can improve individual fairness. Besides, \citet{ruoss2020learning} propose an end-to-end framework to learn individually fair representations with provable certification and demonstrates the transferability of individual fairness using the learned representation. \citet{WickerPW23} further study the certification of distributional individual fairness~\cite{Yurochkin2020Training}, which enforces the individual fairness within a $\gamma$-Wassertein ball of the empirical distribution over a finite set of observed individuals. The proposed distributional individual fairness regularization explicitly enables the transferability of individual fairness under in-the-wild distribution shifts.

\subsection{Transparency}
Transparency helps non-experts understand the decision-making process of a machine learning model and the confidence level of the model in making decisions~\cite{Varshney2022}. For example, interpretability and explainability have recently been studied to enhance transparency, by designing a simpler and more interpretable model~\cite{KohL17,Ribeiro0G16} or providing post-hoc explanations for existing black-box models~\cite{SelvarajuCDVPB17}. As a complementary metric of transparency, uncertainty quantification~\cite{BhattAZLSFMKSTN21} illustrates the prediction confidence of a trained model. As a result, we study two major questions behind transparent transfer learning: what knowledge is being transferred in transfer learning, and how to quantify the uncertainty of transfer learning models.

\subsubsection{Interpretability/Explanability}
\begin{figure}
    \centering
    \includegraphics[width=\linewidth]{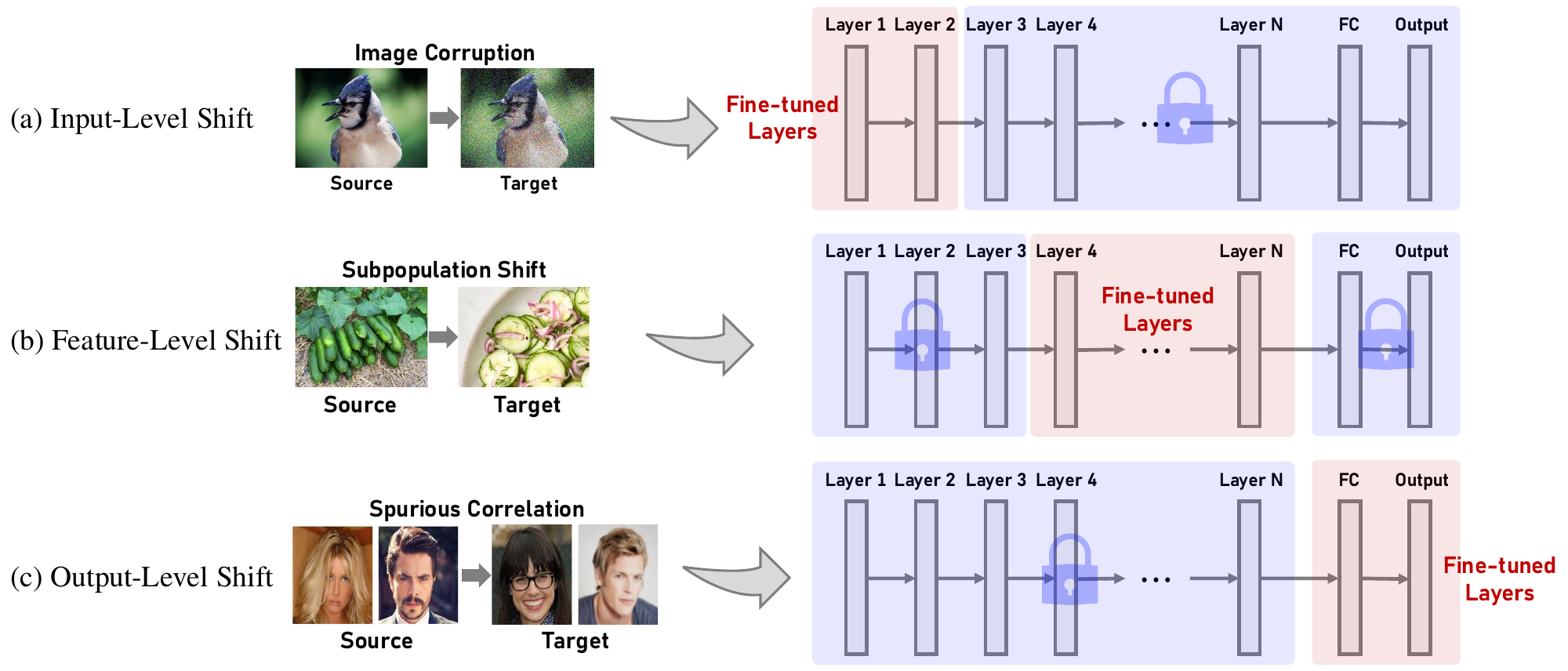}
    \vspace{-5mm}
    \caption{Illustration of surgical fine-tuning (adapted from \cite{lee2023surgical}), where the selected fine-tuning blocks are correlated with the types of distribution shifts between source and target domains.}
    \label{fig:surgicalFT}
    \vspace{-1mm}
\end{figure}

Despite the promising performance of transfer learning techniques in a range of applications, limited effort has been devoted to understanding which data and model architecture components contribute to successful knowledge transfer across domains. To bridge this gap, \citet{YosinskiCBL14} demonstrate that for neural networks pre-trained on ImageNet data set~\cite{deng2009imagenet}, modules in the lower layers are responsible for capturing general features (e.g., Gabor and color blob features in images), while the higher-layer modules tend to encode task-specific semantic features. \citet{NeyshaburSZ20} further support this finding from the perspective of module criticality~\cite{Chatterji2020The}. Besides, they also reveal that both feature reuse and low-level data statistics are crucial for successful knowledge transfer. More recently, \citet{lee2023surgical} establish the connections between fine-tuned neural layers and types of distribution shifts (shown in Fig.~\ref{fig:surgicalFT}). They find that fine-tuning the first block is most effective for input-level shifts (such as image corruption), intermediate blocks excel at feature-level shifts (like shifts in entity subgroups), and tuning the last layer is best for output-level shifts (such as spurious correlations between gender and hair color). \citet{raghu2019transfusion} also investigate transfer learning for medical imaging. They show that using a larger pre-trained ImageNet model does not significantly improve performance compared to smaller lightweight convolutional networks. Additionally, it is observed that transfer learning provides feature-independent benefits, such as improved weight scaling and faster convergence. This is consistent with observations from~\cite{kornblith2019better,he2019rethinking}.

In addition to understanding the transferability of pre-trained models, recent efforts have been devoted to exploring the explanations of distribution shifts across domains in the distribution space. There are two major frameworks for distribution shift explanations, including interpretable transportation mapping~\cite{KulinskiI23,stein2023rectifying} and natural languages~\cite{dunlap2024describing,zhu2022gsclip,zhong2023goal}. Specifically, on one hand, \citet{KulinskiI23,kulinski2022towards} explain distribution shifts using interpretable transportation maps indicating how the source distribution can move to the target distribution in the distribution space. The crucial idea is to leverage optimal transport to find the optimal transportation map from user-defined interpretable candidates. \citet{stein2023rectifying} further propose a group-aware shift explanation framework to rectify the group irregularities when explaining distribution shifts. On the other hand, \citet{zhu2022gsclip} develop a GSCLIP system to explain distribution shifts of different image data sets in natural language. This system generates human-understandable natural language descriptions of distribution shifts as candidate explanations, and then quantitatively evaluates these candidates to identify the most reasonable ones. \citet{zhong2022describing} study the explanations for text distribution shifts through natural languages. They prompt GPT-3~\cite{GPT3} to generate candidate explanations and then employ a verifier neural network to re-rank these explanations. Similarly, \citet{dunlap2024describing} leverage visual-language models to generate candidate difference descriptions from image sets and then re-rank these candidates based on their effectiveness in distinguishing the two sets. In contrast, based on graphical causal models, \citet{budhathoki2021did} propose a Shapley value framework to quantify the attribution for each causal mechanism for distribution shifts. The follow-up work~\cite{zhang2023did} further explores connections between model performance changes across domains and interpretable distribution shifts via Shapley values.


\subsubsection{Uncertainty Quantification}
Uncertainty quantification is essential for decision-making and optimization in machine learning and artificial intelligence~\cite{naeini2015obtaining}. For example, high-stakes applications such as medical diagnostics~\cite{begoli2019need} and autonomous driving~\cite{michelmore2020uncertainty} require both accurate class predictions and quantification of prediction uncertainty. Generally, there are two types of prediction uncertainty~\cite{hullermeier2021aleatoric}: aleatoric (data) uncertainty involving the inherent randomness and variability in the data, and epistemic (model) uncertainty caused by a lack of knowledge about the optimal model parameters. These uncertainties can be formally explained using the Bayesian posterior distribution~\cite{ChanAQS20}.
\begin{definition}[Aleatoric and Epistemic Uncertainty~\cite{ChanAQS20}]
    Given a model $f$ with parameter $\theta$ and a test sample $x^*$, the Bayesian posterior distribution over $x^*$ can be formulated as:
    \begin{align}
       \underbrace{ p(y^* | x^*, D)}_{\text{Total uncertainty}} = \int \underbrace{ p(y^* | x^*, f) }_{\text{Aleatoric uncertainty}} \cdot \underbrace{ p(f | D) }_{\text{Epistemic uncertainty}} d f
    \end{align}
    where $D$ denotes the set of training samples.
\end{definition}

The calibration of uncertainty estimates is vital in determining the trustworthiness of model outputs. A well-calibrated model should provide accurate predictions when it is confident and indicate high uncertainty when it is likely to be incorrect. Thus, calibration can be considered as an orthogonal metric for accuracy when evaluating machine learning systems. In particular, \citet{SnoekOFLNSDRN19} conduct a systematic evaluation of traditional uncertainty quantification models under distribution shifts. They observe that the quality of uncertainty consistently degrades with increasing distribution shifts between source and target domains. To solve this problem, various frameworks have been proposed to improve uncertainty quantification under distribution shifts across domains.

\begin{enumerate}[(1)]
    \item {\em Temperature Scaling:} \citet{ParkBWL20} derive an upper bound on the expected calibration error in the target domain in terms of the importance-weighted classification error and the error of a domain discriminator. Building on the idea of temperature scaling~\cite{guo2017calibration}, they propose a calibration algorithm by minimizing the upper bound over source and target samples. Similarly, \citet{TransCal} develop an adaptive importance weighting approach with lower bias and variance of the estimated calibration errors to improve the uncertainty quantification under the covariate shift assumption. Instead, \citet{zou2023adaptive} focus on learning two calibration functions based on a real in-distribution calibration set and a synthetic out-of-distribution calibration set respectively, and then adaptively combines the two calibrators. \citet{hu2024pseudocalibration} optimize the calibration objective function (i.e., temperature scaling optimization~\cite{guo2017calibration}) using a labeled pseudo-target set created via mixup~\cite{zhang2018mixup} over pseudo-labeled target samples.

    \item {\em  Conformal Prediction:} The model predicts a set of labels instead of a single label~\cite{romano2020classification,lei2018distribution,angelopoulos2021gentle}. Assuming the true importance weights (e.g., $w^*(x,y) = p_T(x,y) / p_S(x,y)$) are known, \citet{tibshirani2019conformal,PodkopaevR21} study the weighted conformal predictions under covariate shifts and label shifts, respectively. Based on jackknife+~\cite{Barber2019PredictiveIW}, \citet{prinster2022jaws,prinster2023jaws} formulate the sampling weighted jackknife+ prediction interval to handle covariate shifts with a finite-sample coverage guarantee. \citet{cauchois2024robust} design prediction sets that are robust against all distribution shifts with bounded $f$-divergence. \citet{gibbs2021adaptive,gibbs2024conformal} further investigate prediction sets in an online setting where the data distribution can shift continuously over time. In addition, \citet{park2022pac} construct probably approximately correct (PAC) prediction sets under bounded covariate shifts in scenarios with known importance weights and an uncertainty set of possible importance weights. The follow-up work~\cite{si2024pac} constructs prediction sets with PAC guarantees in the presence of label shifts. The crucial idea is to compute confidence intervals of importance weights~\cite{lipton2018detecting} through Gaussian elimination.

    \item {\em  Bayesian Learning:} \citet{ChanAQS20} develop an approximate Bayesian inference approach based on posterior regularization that captures the distribution difference between source and target domains.
    \citet{zhou2021bayesian} study the uncertainty quantification problem in the context of test-time adaptation. They develop a probabilistic graphical model for covariate shift scenarios, followed by an instantiated ensemble approach to estimate the uncertainty of trained models over test samples. In addition, transferable Gaussian processes~\cite{yu2007gaussian,bonilla2007multi,cao2010adaptive,maddox2021fast,DINO} can be applied to model uncertainty in the target domain by leveraging knowledge from the source domain.
\end{enumerate}

\subsection{Other Trustworthiness Concerns}

\subsubsection{Accountability and Auditability}
Accountability is crucial to evaluate the trustworthiness of AI outcomes, as it identifies the organizations and individuals responsible for these results. More specifically, \citet{bovens2007analysing} defines accountability as ``{\em a relationship between an actor and a forum, in which the actor has an obligation to explain and to justify his or her conduct, the forum can pose questions and pass judgment, and the actor may face consequences}". \citet{wieringa2020account} further analyze five key aspects of this definition, including the actor, the forum, the relationship between them, the content and criteria of the account, and the potential consequences resulting from the account. In this case, auditability refers to systematic evaluations to guarantee accountability~\cite{raji2020closing}. Given the remarkable performance of fine-tuned large language models (LLMs), auditing LLMs has been studied through different principled assessments~\cite{mokander2023auditing,amirizaniani2024auditllm}. Recently, \citet{pei2023data} discuss data and AI model markets that facilitate the sharing, discovery, and integration of data and AI models among multiple parties. These markets can enhance knowledge transfer between pre-trained AI models and user-specific tasks, but they raise fundamental concerns regarding accountability in these systems. Further research can be conducted to guarantee accountability for transfer learning systems in supporting model and data knowledge sharing.

\subsubsection{Sustainability and Environmental Well-being}
To establish the trustworthiness of machine learning and artificial intelligence systems, it is crucial to evaluate resource usage and energy consumption within their entire supply chain~\cite{nikolinakos2023ethical,budennyy2022eco2ai}. Notably, \citet{schwartz2020green} introduce a simple notion of computational cost in producing AI results.
\begin{definition}[Cost of an AI Result~\cite{schwartz2020green}]
    The total cost of producing a (R)esult in AI increases with the following quantities.
    \begin{equation}
        \mathrm{Cost}(R) \propto E \cdot D \cdot H
    \end{equation}
    where $E$ is the cost of executing the model on a single (E)xample, $D$ is the size of the training (D)ataset, and $H$ is the number of (H)yperparameter experiments.
\end{definition}
To reduce the computational cost, green AI~\cite{schwartz2020green,huang2024towards,memmel2024position} has been promoted by improving the efficiency of AI models with positive impacts on the environment. Several efficiency metrics have been introduced, including carbon emission, electricity usage, floating-point operations (FLOPs), elapsed runtime, and the number of parameters. Transfer learning techniques demonstrated significant improvements in training efficiency by leveraging knowledge from pre-trained models~\cite{YosinskiCBL14,he2019rethinking}. This is because these approaches reduce (1) the size of training data $D$ and hyperparameters $H$, and (2) the number of trainable model parameters (via parameter-efficient fine-tuning~\cite{Adapters,LoRA}). Furthermore, \citet{huang2024towards} recently propose a GreenTrainer method to minimize the FLOPs of LLM fine-tuning via adaptive backpropagation. \citet{qiu2023first} take a first look into the carbon footprint of federated learning, by quantifying carbon emissions from hardware training and communication between server and clients. In real-world applications, transfer learning has been applied to lower energy consumption and reduce carbon emissions by reusing pre-trained models. For example, \citet{kunwar2024managing,ahmed2024flowers} analyze transfer learning techniques for garbage classification and flower classification in terms of both prediction accuracy and carbon emissions.

%% file: content/5_applications.tex
\section{Applications}\label{sec:applications}
Trustworthy transfer learning has been widely applied to artificial intelligence and machine learning fields, including computer vision~\cite{NeyshaburSZ20}, natural language processing~\cite{ding2023parameter}, and graph learning~\cite{WNN}. In addition, this section highlights real-world applications of trustworthy transfer learning in scientific discovery.

\subsection{Agriculture}
Transfer learning techniques have been applied to various precision agriculture applications~\cite{ma2024transfer,adve2024aifarms}. Specifically, to improve the management of agricultural stakeholders, \citet{zhang2021transfer,wang2023airborne} propose process-guided machine learning frameworks, which transfer knowledge from simulated data generated by soil-vegetation radiative transfer modeling to real-world field data for precise monitoring of cover crop traits. \citet{wan2022combining} analyze the transferability of support vector regression models for estimating leaf nitrogen concentration across different plant species. Besides, pre-trained vision models have been fine-tuned for crop mapping~\cite{jo2022towards}, crop pest classification~\cite{thenmozhi2019crop}, and plant phenotyping~\cite{sama2023new}.

\subsection{Bioinformatics}
Notably, \citet{Theodoris2023TransferLE} introduce an attention-based foundation model Geneformer pre-trained on over 30 million single-cell transcriptomes to capture network dynamics (e.g., gene interactions). They also demonstrate the effectiveness of Geneformer in various downstream tasks with limited data through fine-tuning. Later, \citet{hou2024assessing} illustrate the efficacy of the pre-trained large language model GPT-4 in cell type annotation of single-cell RNA-seq data. Besides, \citet{hu2020iterative} develop a unified transfer learning framework for open-world single-cell classification across different species and tissues. Similarly, \citet{mieth2019using} study the clustering of single-cell RNA-seq data on the small disease- or tissue-specific data sets by leveraging prior knowledge from large reference data sets. \citet{hetzel2022predicting,lotfollahi2022mapping} leverage architecture surgery based transfer learning techniques to understand cellular heterogeneity.

In addition, recent efforts~\cite{rao2019evaluating,detlefsen2022learning,heinzinger2019modeling} have been devoted to protein representation learning for downstream tasks using language models pre-trained on a large protein corpus. Typically, \citet{rao2019evaluating} introduce a protein transfer learning benchmark TAPE for learning transferable protein representation, and \citet{detlefsen2022learning} further improve the quality of protein representation by considering the geometry of representation space. \citet{dieckhaus2024transfer} also exploit the pre-trained ProteinMPNN model~\cite{dauparas2022robust} to extract embeddings of input proteins, which are then used to predict stability changes for protein point mutations.

\subsection{Healthcare}
Transfer learning advances the development of effective and efficient health care services~\cite{jayaraman2020healthcare}. For example, \citet{FedHealth} develop a federated transfer learning framework for privacy-preserving wearable healthcare systems (e.g., Parkinson’s disease auxiliary diagnosis). \citet{raghu2019transfusion,matsoukas2022makes} further understand the impact of the source domain/model on the downstream medical imaging tasks in the context of transfer learning. To enforce health equity across ethnic groups, \citet{gao2020deep,toseef2022reducing,lee2023reducing} propose transferring knowledge from majority groups with sufficient data to minority groups with limited data. In addition, transfer learning techniques have been applied to drug discovery~\cite{cai2020transfer}. Specifically, \citet{yao2021functionally} propose a functional rationalized meta-learning algorithm to enable knowledge transfer across assays for virtual screening and ADMET prediction. \citet{goh2018using,dalkiran2023transfer} adopt pre-training and fine-tuning strategies for molecular property prediction and drug-target interaction prediction, respectively.

\subsection{Education}
Transfer learning has been studied in Educational Data Mining (EDM) for predicting student performance in higher education~\cite{hunt2017transfer}. Over the past decade, Massive Open Online Courses (MOOCs) have supported millions of learners around the world. Early predictions of student performance are crucial for enabling timely interventions in these courses. Transfer learning has been explored to predict student performance in ongoing courses by leveraging knowledge from previous courses. To be specific, \citet{boyer2015transfer} leverage knowledge from both previous courses and previous weeks of the same course to make real-time predictions for learners in MOOCs. Instead of relying on handcrafted features, \citet{ding2019transfer} aim to learn domain-invariant representation by using an auto-encoder and correlation alignment~\cite{CORAL} between source and target courses. Similarly, \citet{SwamyMK22} study the transferability of early success prediction models across MOOCs from different domains and topics. Besides, \citet{Schmucker2022transferable} explore the transferability of student performance in addressing the cold-start problem for new courses in intelligent tutoring systems. More recently, large language models such as GPT-4 and ChatGPT have gained significant attention for improving instructional efficiency and student engagement (e.g., by creating interactive homework with feedback and follow-up questions)~\cite{vanzo2024gpt,Kasneci2023ChatGPTFG}. 

\subsection{Robotics}\label{sec:robotics}
Sim-to-real transfer aims to transfer knowledge from simulation to real-world environments when training reinforcement learning models for robotic learning~\cite{dai2024acdc,peng2018sim}. Recently, this framework has been studied in various robotic learning tasks, including Rubik’s cube~\cite{akkaya2019solving}, human pose estimation~\cite{DoerschZ19}, vision-and-language navigation~\cite{AndersonSTMPBL20}, biped locomotion~\cite{YuKTL19}, etc. Specifically, various strategies have been proposed to address the domain shift between simulation and real-world environments~\cite{tzeng2020adapting,peng2018sim,pinto2017robust,rusu2017sim}. One is to use distribution alignment regularization to learn domain-invariant representation~\cite{tzeng2020adapting,DIRL}. Another strategy is domain randomization~\cite{andrychowicz2020learning,tobin2017domain}, which aims to train the model using a diverse set of randomized simulated environments, rather than relying on a single simulated environment. \citet{chen2022understanding,hu2023provable} further theoretically highlight the benefits of domain randomization for sim-to-real transfer. 

\subsection{E-commerce}
Cross-domain recommendation aims to generate reliable recommendations in a target domain by exploiting knowledge from source recommender systems. It has been studied from various perspectives, e.g., matrix factorization~\cite{man2017cross,samra2024cross}, neural collaborative filtering~\cite{HuZY18,kanagawa2019cross,li2020ddtcdr}, graph neural network~\cite{GRADE,zhao2019cross}, large language models~\cite{petruzzelli2024instructing}, etc. In addition to prediction accuracy, recent works also investigate the trustworthiness properties of cross-domain recommender systems, such as adversarial vulnerability~\cite{chen2019data} and user privacy~\cite{yang2024federated}. 




%% file: content/6_future.tex
\section{Open Questions and Future Trends}\label{sec:future}
Despite the rapidly increasing research interest and applications of trustworthy transfer learning in both academia and industry, there remain many open questions, especially in the theoretical understanding of trustworthy transfer learning. 

\subsection{Benchmarking Negative Transfer}
Negative transfer can roughly be defined as the phenomenon~\cite{TL_survey_2010} where ``{\em transferring knowledge from the source can have a negative impact on the target learner}".
\begin{definition}[Negative Transfer~\cite{TL_survey_2010,WangDPC19}]
    Given a learning algorithm $A^{\mathrm{tl}}$, source data $D_S$, and target data $D_T$, negative transfer occurs if the following condition holds:
    \begin{align}
    \mathcal{E}_{T} \left( A^{\mathrm{tl}}\left( D_S, D_T \right) \right) > \mathcal{E}_{T} \left( A^{\mathrm{tl}}\left( \emptyset, D_T \right) \right)
    \end{align}
    where $\mathcal{E}_{T} ( A^{\mathrm{tl}}\left( S \right) )$ represents the expected error on the target distribution $\mathrm{P}_T$ when the learning algorithm $A^{\mathrm{tl}}$ is trained on data $S$.
\end{definition}
This definition reveals~\cite{A_distance,KuzborskijO13,WangDPC19} that given a learning algorithm, there are two major factors determining if negative transfer occurs: the {\em distribution discrepancy} between source and target domains and the {\em size of the labeled target data}. Negative transfer has been observed theoretically and empirically in various applications~\cite{rosenstein2005transfer,Ben-DavidLLP10,0002CZG19,WangDPC19}. Recent work~\cite{cohen2024ask} also explores identifying and characterizing the failure modes that pre-training can and cannot address. Despite the extensive work on transfer learning techniques, up until now, little effort (if any) has been devoted to rigorously understanding the boundary between positive and negative transfer given a learning algorithm. It remains an open question to determine when the negative transfer will occur given finite source and target samples (or a source model and finite target samples). Therefore, rather than focusing on performance improvement, more efforts can be dedicated to benchmarking the negative transfer of transfer learning models, e.g., the change from positive to negative transfer can be affected by the magnitude of distribution shifts and the number of target samples. This could provide valuable insights into when a transfer learning model can work well for real-world applications.

\subsection{Cross-modal Transferability}
Cross-modal transfer learning~\cite{ShenLDSKNT23,DinhZZLGRSP022,socher2013zero} aims at understanding knowledge transferability when the source and target domains have different types of data modalities, e.g., transferring knowledge from a text-based source domain to an image-based target domain. This differs from multi-modal learning~\cite{RadfordKHRGASAM21,huang2021makes}, which maps different data modalities in a unified latent feature space over pair-wise training samples. In contrast, cross-modal transfer learning focuses on investigating what knowledge can be transferred across data modalities. Although large language models (LLMs) have been applied to various scientific discovery tasks~\cite{DinhZZLGRSP022}, it is unclear what knowledge is being transferred in this process. Additionally, there is a lack of theoretical understanding regarding how LLMs generalize to downstream tasks with different data modalities.

\subsection{Physics-informed Transfer Learning}
Physics-informed machine learning~\cite{karniadakis2021physics,raissi2019physics} aims to improve the training of machine learning models by incorporating physical domain knowledge as soft constraints on an empirical loss function. This alleviates the need for a large amount of high-quality data when training deep neural networks to solve scientific problems. Recent studies have introduced transfer learning to understand the knowledge transferability of physics-informed neural networks across tasks. For example, \citet{desai2022oneshot} study the transferability of physics-informed neural networks across differential equations. \citet{goswami2022deep,xu2023transfer} study the transfer learning performance of deep operator networks across partial differential equations. \citet{subramanian2023towards} further analyze the transfer behavior of neural operators pre-trained on a mixture of different physics problems. However, the theoretical explanation regarding the generalization performance of physics-informed neural networks under distribution shifts is unclear. Besides, it can be seen that in the context of transfer learning, the source knowledge can be provided from multiple aspects, including labeled source samples~\cite{wiles2022a} (e.g., Subsection~\ref{sec:distribution_discrepancy}), pre-trained source models~\cite{GPT4} (e.g., Subsection~\ref{sec:htl}), synthetic data generated by physics-based simulators~\cite{andrychowicz2020learning} (e.g., Subsection~\ref{sec:robotics}), and fundamental physical rules~\cite{karniadakis2021physics}. This can motivate a generic physics-informed transfer learning problem involving multi-faceted knowledge transfer from the source to the target domains.

\subsection{Trade-off between Transferability and Trustworthiness}
In standard machine learning, the trade-off between prediction accuracy and trustworthiness has been theoretically studied, e.g., accuracy vs. group fairness~\cite{ZhaoG19,DuttaWYC0V20}, accuracy vs. adversarial robustness~\cite{TsiprasSETM19,ZhangYJXGJ19,YangRZSC20}, accuracy vs. privacy~\cite{BiettiWD0W22}, accuracy vs. explainability~\cite{ZarlengaBCMGDSP22}, etc. It has been observed that trustworthiness may be at odds with the prediction accuracy in a single domain. In contrast, recent work~\cite{SalmanIEKM20,davchev2019empirical} reveals that both trustworthiness (e.g., adversarial robustness) and prediction accuracy can be improved in the target domain by leveraging relevant knowledge from source domains. This motivates us to rethink the fundamental trade-off between knowledge transferability and trustworthiness in the context of transfer learning. Specifically, there are several open questions: (1) Can source knowledge consistently enhance trustworthiness and transfer accuracy in the target domain under various distribution shifts and data modalities? (2) Is there an inherent trade-off between trustworthiness and transfer accuracy in the target domain when discovering and transferring knowledge from the source data/model? These studies will significantly expand the application of transfer learning techniques by clarifying when trained models can be trusted and how well they perform.

\subsection{Trustworthy Transfer Learning of Foundation Models}
Although foundation models have achieved surprising performance across high-impact domains, the underlying properties governing their behavior are not yet fully understood~\cite{bengio2024managing,kapoor2024position}, especially when adapting them to downstream tasks. In this work, we summarize the key open questions from the perspective of trustworthy transfer learning. On one hand, knowledge transferability involves whether foundation models can be successfully adapted to downstream tasks with guaranteed performance. This post-training process leads to the following open questions:
\begin{itemize}
    \item {\em Scaling Behaviors of Fine-tuning:} 
    Recent work~\cite{tay2022scale,zhang2024when} has investigated the power-law scaling relationship between LLM fine-tuning performance and model/data size. In contrast, \citet{ren2025learning} provide a step-wise analysis of the learning dynamics for LLM fine-tuning under gradient descent. However, theoretical understanding still lags behind empirical findings, e.g., how generalization performance of fine-tuning is bounded in terms of model architecture, model and data scale, how learning dynamics vary across different fine-tuning (or other post-training) strategies, and how learning dynamics of fine-tuning explain the potential negative transfer behaviors.

    \item {\em Transferability Metrics for Foundation Model Selection:}
    Recent work~\cite{lin2024selecting} leverages rectified scaling laws to predict fine-tuning performance, thereby facilitating model selection for LLMs. However, it is computationally expensive due to the need for fine-tuning in low-data regimes, and the theoretical foundations connecting scaling laws to fine-tuning performance are still underexplored. To enable model selection and customization in the AI model market~\cite{pei2023data}, there is a growing need for theory-grounded and computationally efficient transferability metrics. In particular, it is feasible and desirable to establish principled guidelines for selecting optimal foundation models for specific target tasks, as well as identifying the application domains in which a given foundation model is best suited for fine-tuning.

    \item {\em AI Model Collapse:} 
    Generally, model collapse is defined as ``a degenerative process affecting generations of learned generative models, in which the data they generate end up polluting the training set of the next generation"~\cite{shumailov2024ai}. This phenomenon motivates us to rethink the impact of synthetic data (maybe indistinguishable from human-generated content) on both the pre-training and fine-tuning of foundation models. From the perspective of transfer learning, a key challenge for mitigating model collapse lies in handling open-world distribution shifts between real and AI-generated data.
    
\end{itemize}
On the other hand, the knowledge trustworthiness of foundation models in the transfer learning process might involve the following open questions:
\begin{itemize}
    \item {\em Data Poisoning and Model Stealing Attacks:}
    It is shown~\cite{WanWSK23,bowen2024scaling} that larger foundation models are significantly more susceptible to data poisoning attacks than smaller ones during fine-tuning. However, limited efforts have been devoted to theoretically understanding the connections between model size and adversarial vulnerability of foundation models in the context of transfer learning. In addition, recent work~\cite{carlini2024stealing} discusses the model stealing attacks to extract precise information from black-box production language models via APIs. These emerging threats highlight significant security challenges for the development and deployment of foundation models in both academia and industry. As a result, advancing certified robust techniques is essential to improve the reliability and trustworthiness of foundation models.

    \item {\em Transparency and Mechanistic Interpretability.} Mechanistic interpretability aims at understanding the internal mechanisms of foundation models~\cite{dunefsky2024transcoders,ferrando2025do}, thereby providing insights into their decision-making processes and enhancing model transparency. Nevertheless, due to the architectural complexity and scale of foundation models, it is challenging to systematically explain their emergent behaviors and key mechanisms behind their advanced capabilities, when they are adapted to open-world environments. Although fragmented efforts have been made, a systematic framework is still needed to advance the mechanistic interpretability of foundation models.
    
    \item {\em Ethical and Societal Impact in Foundation Model Customization:} The ethical and societal implications, such as misinformation and privacy concerns, of open foundation models have been examined from the perspectives of AI developers~\cite{nikolinakos2023ethical,kapoor2024position}. As these models are increasingly fine-tuned and customized by millions of non-expert users, a critical gap emerges between expert understanding and everyday use. This gap highlights the urgent need for intuitive risk assessments and protective measures that empower non-AI users to identify and mitigate intentional or unintentional misuse of foundation models.    
\end{itemize}

%% file: content/7_conclusion.tex
\section{Conclusion}\label{sec:conclusion}
In this survey, we provide a comprehensive review of trustworthy transfer learning from the perspective of knowledge transferability and trustworthiness. With different data and model assumptions, much effort has been devoted to understanding the generalization performance of trustworthy transfer learning and designing advanced techniques in quantifying and enhancing knowledge transfer in a variety of real-world applications. Besides, we also summarize several open questions for trustworthy transfer learning, including benchmarking positive and negative transfer, enabling unified knowledge transfer across different data modalities and physical rules, and achieving the inherent trade-off between transferability and trustworthiness. Ultimately, trustworthy transfer learning could lead to a unified machine learning and artificial intelligence framework that facilitates positive knowledge reuse and transfer in the presence of distribution shifts and across data modalities, while maintaining rigorous standards of trustworthiness.